\documentclass[10pt,twocolumn,letterpaper]{article}

\usepackage[pagenumbers]{cvpr} % To force page numbers, \eg for an arXiv version

\usepackage{url}
\usepackage{graphicx}
\usepackage{amsmath}
\usepackage{amsthm}
\usepackage{amsfonts,bm}
\usepackage{amssymb}
\usepackage{multirow}
\usepackage{booktabs}
\newtheorem{theorem}{Theorem}
\newtheorem{definition}{Definition}
\newtheorem{lemma}{Lemma}
\newtheorem{assumption}{Assumption}
\usepackage{subfloat}
\usepackage{tabularx}
\usepackage{color}
\usepackage[pagebackref,breaklinks,colorlinks]{hyperref}

\usepackage[capitalize]{cleveref}
\crefname{section}{Sec.}{Secs.}
\Crefname{section}{Section}{Sections}
\Crefname{align}{Equation}{Equations}
\crefname{align}{Eq.}{Eqs.}
\Crefname{equation}{Equation}{Equations}
\crefname{equation}{Eq.}{Eqs.}
\Crefname{figure}{Figure}{Figures}
\crefname{figure}{Fig.}{Figs.}
\crefname{assumption}{Assumption}{Assumptions}

\def\cvprPaperID{7717} % *** Enter the CVPR Paper ID here
\def\confName{CVPR}
\def\confYear{2022}

\begin{document}

\title{Rethinking Minimal Sufficient Representation in Contrastive Learning}
% What Makes for Good Representations for Contrastive Learning

% \author{
% Haoqing Wang\thanks{The work was done when the author was with MSRA as an intern.}\\
% Peking University\\
% \footnotesize \href{mailto:wanghaoqing@pku.edu.cn}{wanghaoqing@pku.edu.cn}\\
% \and
% Xun Guo\\
% Microsoft Research Asia\\
% \footnotesize \href{mailto:xunguo@microsoft.com}{xunguo@microsoft.com}\\
% \and
% Zhi-Hong Deng\thanks{Corresponding author.}\\
% Peking University\\
% \footnotesize \href{mailto:zhdeng@pku.edu.cn}{zhdeng@pku.edu.cn}\\
% \and
% Yan Lu\\
% Microsoft Research Asia\\
% \footnotesize \href{mailto:yanlu@microsoft.com}{yanlu@microsoft.com}
% }

\author{
Haoqing Wang\thanks{The work was done when the author was with MSRA as an intern.} $^1\qquad$  Xun Guo$^2\qquad$  Zhi-Hong Deng$^1\qquad$  Yan Lu$^2$ \\
$^1$Peking University  $\qquad^2$Microsoft Research Asia \\
\small \texttt{$\{$wanghaoqing, zhdeng$\}$@pku.edu.cn} $\qquad$ \texttt{$\{$xunguo, yanlu$\}$@microsoft.com} \\
}
\maketitle

%%%%%%%%% ABSTRACT
\begin{abstract}
Contrastive learning between different views of the data achieves outstanding success in the field of self-supervised representation learning and the learned representations are useful in broad downstream tasks. Since all supervision information for one view comes from the other view, contrastive learning approximately obtains the minimal sufficient representation which contains the shared information and eliminates the non-shared information between views. Considering the diversity of the downstream tasks, it cannot be guaranteed that all task-relevant information is shared between views. Therefore, we assume the non-shared task-relevant information cannot be ignored and theoretically prove that the minimal sufficient representation in contrastive learning is not sufficient for the downstream tasks, which causes performance degradation. This reveals a new problem that the contrastive learning models have the risk of over-fitting to the shared information between views. To alleviate this problem, we propose to increase the mutual information between the representation and input as regularization to approximately introduce more task-relevant information, since we cannot utilize any downstream task information during training. Extensive experiments verify the rationality of our analysis and the effectiveness of our method. It significantly improves the performance of several classic contrastive learning models in downstream tasks. Our code is available at \url{https://github.com/Haoqing-Wang/InfoCL}.
\end{abstract}

%%%%%%%%% BODY TEXT
\section{Introduction}
Recently, contrastive learning \cite{chen2020simple,grill2020bootstrap,DBLP:conf/nips/CaronMMGBJ20,DBLP:conf/icml/ZbontarJMLD21,DBLP:conf/cvpr/ChenH21} between different views of the data achieves outstanding success in the field of self-supervised representation learning. The learned representations are useful for broad downstream tasks in practice, such as classification, detection and segmentation \cite{he2020momentum}.  In contrastive learning, the representation that contains all shared information between views is defined as \emph{sufficient representation}, while the representation that contains only the shared and eliminates the non-shared information is defined as \emph{minimal sufficient representation} \cite{DBLP:conf/iclr/Tsai0SM21}. Contrastive learning maximizes the mutual information between the representations of different views, thereby obtaining the sufficient representation. Furthermore, since all supervision information for one view comes from the other view \cite{DBLP:conf/iclr/Federici0FKA20}, the non-shared information is often ignored, so that the minimal sufficient representation is approximately obtained.

\begin{figure}
\centering
\includegraphics[width=1.\linewidth]{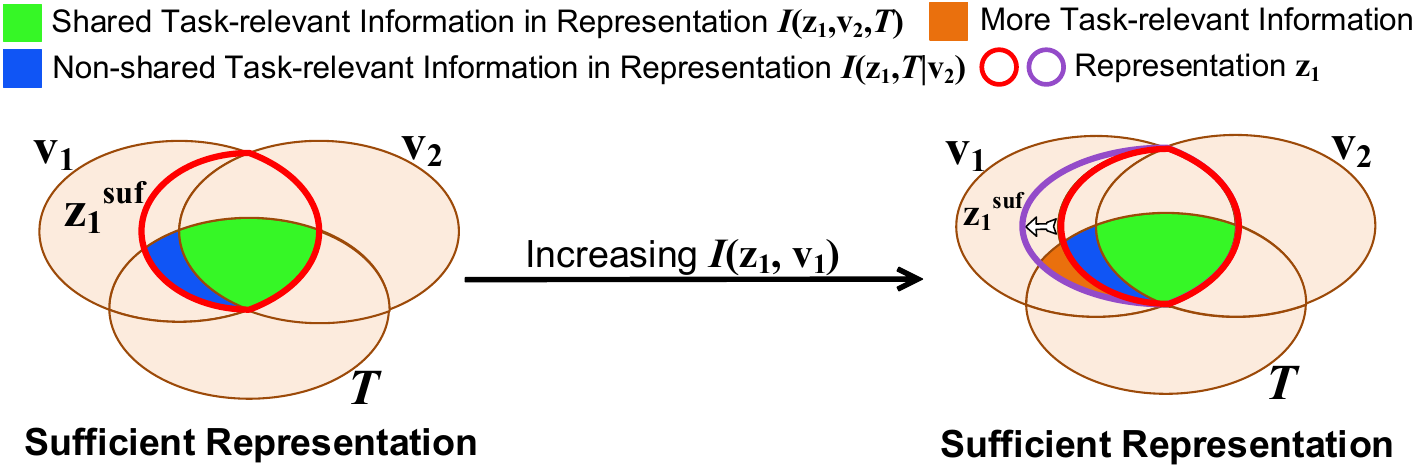}
\caption{Demonstration of our motivation using information diagrams. Based on the (approximately minimal) sufficient representation learned in contrastive learning, increasing $I(z_1,v_1)$ approximately introduces more non-shared task-relevant information.}
\label{motivation}
\end{figure}

Tian \etal \cite{DBLP:conf/nips/Tian0PKSI20} find that the optimal views for contrastive learning depend on the downstream tasks when the minimal sufficient representation is obtained. In other words, the optimal views for task $T_1$ may not be suitable for task $T_2$. The reason may be that some information relevant to $T_2$ is not shared between these views. In this work, we formalize this conjecture and assume that the non-shared task-relevant information cannot be ignored. Based on this assumption, we theoretically prove that the minimal sufficient representation contains less task-relevant information than other sufficient representations and has a non-ignorable gap with the optimal representation, which causes performance degradation. Concretely, we consider two types of the downstream task, \ie, classification and regression task, and prove that the lowest achievable error of the minimal sufficient representation is higher than other sufficient representations.

According to our analysis, when some task-relevant information is not shared between views, the learned representation in contrastive learning is not sufficient for the downstream tasks. This reveals that the contrastive learning models have the risk of over-fitting to the shared information between views. To this end, we need to introduce more non-shared task-relevant information to the representations. Since we cannot utilize any downstream task information in the training stage, it is impossible to achieve this directly. As an alternative, we propose an objective term which increases the mutual information between the representation and input to approximately introduce more task-relevant information. This motivation is demonstrated in \cref{motivation} using information diagrams. We consider two implementations to increase the mutual information. The first one reconstructs the input to make the representations containing the key information about the input \cite{DBLP:journals/jmlr/VincentLLBM10,DBLP:journals/corr/KingmaW13}. The second one relies on the high-dimensional mutual information estimate \cite{belghazi2018mutual,poole2019variational}.

Overall, we summarize our contributions as follows.
\begin{itemize}
\item To the best of our knowledge, this is the first work to theoretically reveal that contrastive learning has the risk of over-fitting to the shared information between views. We provide comprehensive analysis based on the internal mechanism of contrastive learning that the views provide supervision information to each other.
\item To alleviate this problem, when the downstream task information is not available, we propose to increase the mutual information between the representation and input to approximately introduce more task-relevant information, as shown in \cref{motivation}.
\item We verify the effectiveness of our method for SimCLR \cite{chen2020simple}, BYOL \cite{grill2020bootstrap} and Barlow Twins \cite{DBLP:conf/icml/ZbontarJMLD21} in classification, detection and segmentation tasks. We also provide extensive analytical experiments to further understand our hypotheses, theoretical analysis and model.
\end{itemize}

\section{Related works}
\paragraph{Contrastive learning.} Contrastive learning between different views of the data is a successful self-supervised representation learning framework. The views are constructed by exploiting the structure of the unlabeled data, such as local patches and the whole image \cite{hjelm2018learning}, different augmentations of the same image \cite{wu2018unsupervised,bachman2019learning,he2020momentum,chen2020simple}, or video and text pairs \cite{sun2019learning,miech2020end}. Recently, Tian \etal\cite{DBLP:conf/nips/Tian0PKSI20} find that the optimal views for contrastive learning are task-dependent under the assumption of minimal sufficient representation. In other words, even if the given views are optimal for some downstream tasks, they may not be suitable for other tasks. In this work, we theoretically analyze this discovery and find that the contrastive learning models may over-fit to the shared information between views, and thus propose to increase the mutual information between representation and input to alleviate this problem. Some recent works \cite{DBLP:conf/iclr/Federici0FKA20,DBLP:conf/iclr/Tsai0SM21} propose to learn the minimal sufficient representation. They assume that almost all the information relevant to downstream tasks is shared between views, which is an overly idealistic assumption and conflicts with the discovery in \cite{DBLP:conf/nips/Tian0PKSI20}.

\paragraph{Information bottleneck theory.} Based on the information bottleneck theory \cite{tishby2015deep,shwartz2017opening,tishby2000information}, a model extracts all task-relevant information in the first phase of learning (drift phase) to ensure sufficiency, and then compresses the task-irrelevant information in the second phase (diffusion phase). Our analysis shows that the learned representation in contrastive learning is not sufficient for the downstream tasks and can be seen as in the drift phase. We need to introduce more task-relevant information to achieve sufficiency.

\section{Theoretical analysis and model}
In this section, we first introduce the contrastive learning framework and theoretically analyze the disadvantages of minimal sufficient representation in contrastive learning, and then propose our method to approximately introduce more task-relevant information to the representations. Note that although the analysis is about the information content in the representations, its specific form is also very important. Therefore, our theoretical analysis is actually based on the premise that the information content in the representations is represented in the most appropriate form.

\subsection{Contrastive learning}

\begin{figure}
\centering
\includegraphics[width=1.\linewidth]{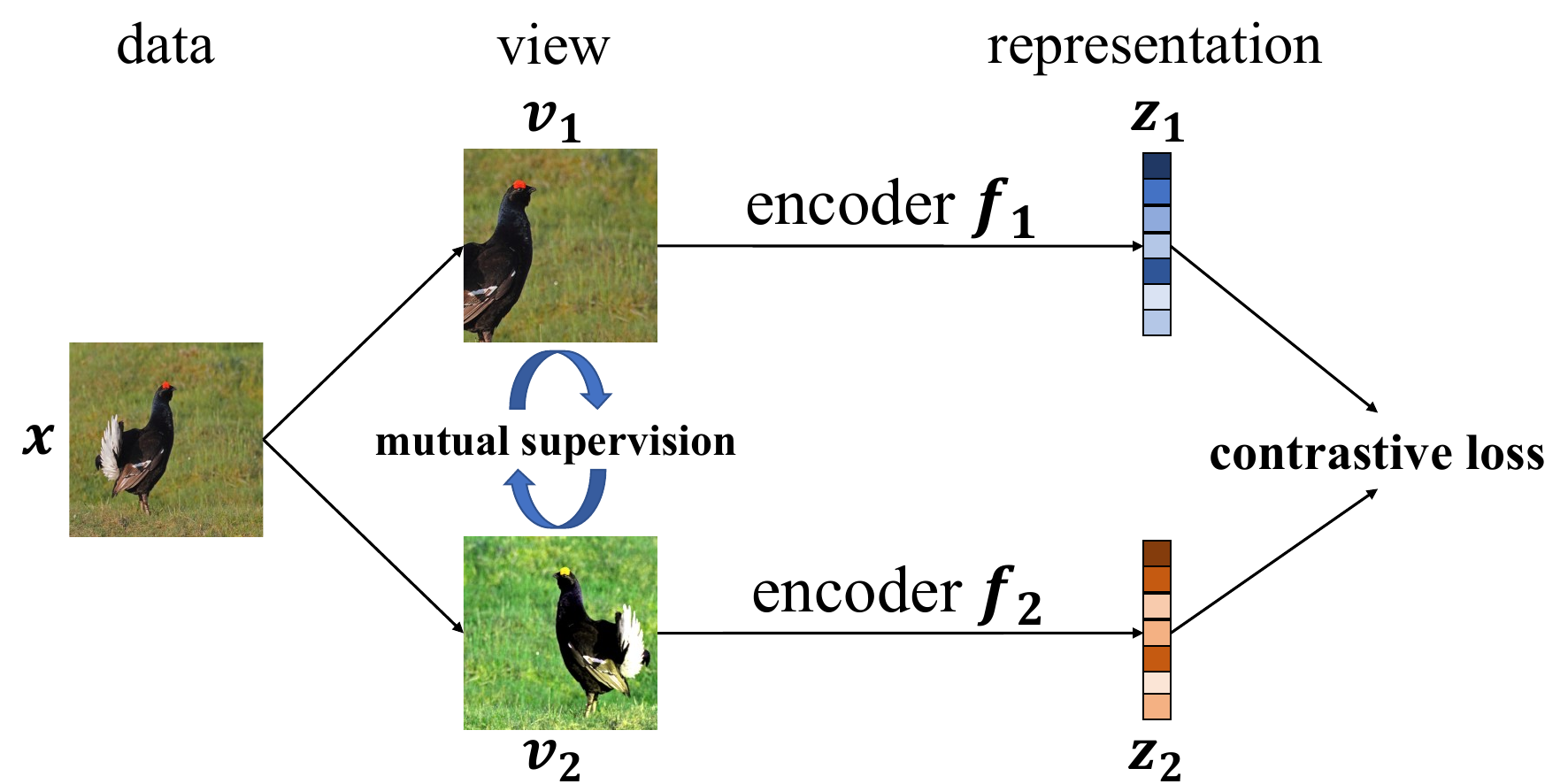}
\caption{Internal mechanism of contrastive learning: the views provide supervision information to each other.}
\label{CL_fig}
\end{figure}

Contrastive learning is a general framework for unsupervised representation learning which maximizes the mutual information between the representations of two random variables $v_1$ and $v_2$ with the joint distribution $p(v_1,v_2)$
\begin{equation}\label{orig}
    \max_{f_1,f_2}I(z_1,z_2)
\end{equation}
where $z_i=f_i(v_i),i=1,2$ are also random variables and $f_i,i=1,2$ are encoding functions. In practice, $v_1$ and $v_2$ are usually two views of the data $x$. When $v_1$ and $v_2$ have the same marginal distributions ($p(v_1)=p(v_2)$), the function $f_1$ and $f_2$ can be the same ($f_1=f_2$).

In contrastive learning, the variable $v_2$ provides supervision information for $v_1$ and plays the similar role as the label $y$ in the supervised learning, and vice versa \cite{DBLP:conf/iclr/Federici0FKA20}. This internal mechanism is illustrated in \cref{CL_fig}. Similar to the information bottleneck theory \cite{tishby2015deep,achille2018emergence} in the supervised learning, we can define the sufficient representation and minimal sufficient representation of $v_1$ (or $v_2$) for $v_2$ (or $v_1$) in contrastive learning \cite{DBLP:conf/nips/Tian0PKSI20,DBLP:conf/iclr/Tsai0SM21}.
\begin{definition}\label{def1}
(Sufficient Representation in Contrastive Learning) The representation $z_1^{suf}$ of $v_1$ is sufficient for $v_2$ if and only if $I(z_1^{suf},v_2)=I(v_1,v_2)$.
\end{definition}
The sufficient representation $z_1^{suf}$ of $v_1$ keeps all the information about $v_2$ in $v_1$. In other words, $z_1^{suf}$ contains all the shared information between $v_1$ and $v_2$, \ie, $I(v_1,v_2|z_1^{suf})=0$. Symmetrically, the sufficient representation $z_2^{suf}$ of $v_2$ for $v_1$ satisfies $I(v_1,z_2^{suf})=I(v_1,v_2)$.
\begin{definition}\label{def2}
(Minimal Sufficient Representation in Contrastive Learning) The sufficient representation $z_1^{min}$ of $v_1$ is minimal if and only if $I(z_1^{min},v_1)\leq I(z_1^{suf},v_1)$, $\forall z_1^{suf}$ that is sufficient.
\end{definition}
Among all sufficient representations, the minimal sufficient representation $z_1^{min}$ contains the least information about $v_1$. Further, it is usually assumed that $z_1^{min}$ only contains the shared information between views and eliminates other non-shared information, \ie, $I(z_1^{min},v_1|v_2)=0$.
% Note that for a specific input instance, the representation can extract the patterns corresponding to its information from this input to obtain the feature of the instance.

Applying Data Processing Inequality \cite{DBLP:books/daglib/0016881} to the Markov chain $v_1\rightarrow v_2\rightarrow z_2$ and $z_2\rightarrow v_1\rightarrow z_1$, we have
\begin{equation}
    I(v_1,v_2)\geq I(v_1,z_2)\geq I(z_1,z_2)
\end{equation}
\ie, $I(v_1,v_2)$ is the upper bound of $I(z_1,z_2)$. Considering that $I(v_1,v_2)$ remains unchanged during the optimization process, contrastive learning optimizes the functions $f_1$ and $f_2$ so that $I(z_1,z_2)$ approximates $I(v_1,v_2)$. When these functions have enough capacity and are well learned based on sufficient data, we can assume $I(z_1,z_2)=I(v_1,v_2)$, which means the learned representations in contrastive learning are sufficient. They are also approximately minimal since all supervision information comes from the other view. Therefore, the shared information controls the properties of the representations.

The learned representations in contrastive learning are typically used in various downstream tasks, so we introduce a random variable $T$ to represent the information required for a downstream task which can be classification, regression or clustering task. Tian \etal \cite{DBLP:conf/nips/Tian0PKSI20} find that the optimal views for contrastive learning are task-dependent under the assumption of minimal sufficient representation. This discovery is intuitive since various downstream tasks need different information that is unknown during training. It is difficult for the given views to share all the information required by these tasks. For example, when one view is a video stream and the other view is an audio stream, the shared information is sufficient for identity recognition task, but not for object tracking task. Some task-relevant information may not lie in the shared information between views, \ie, $I(v_1,T|v_2)$ cannot be ignored. Eliminating all non-shared information has the risk of damaging the performance of the representations in the downstream tasks.

\subsection{Analysis on minimal sufficient representation}

\begin{figure}
\centering
\includegraphics[width=1.\linewidth]{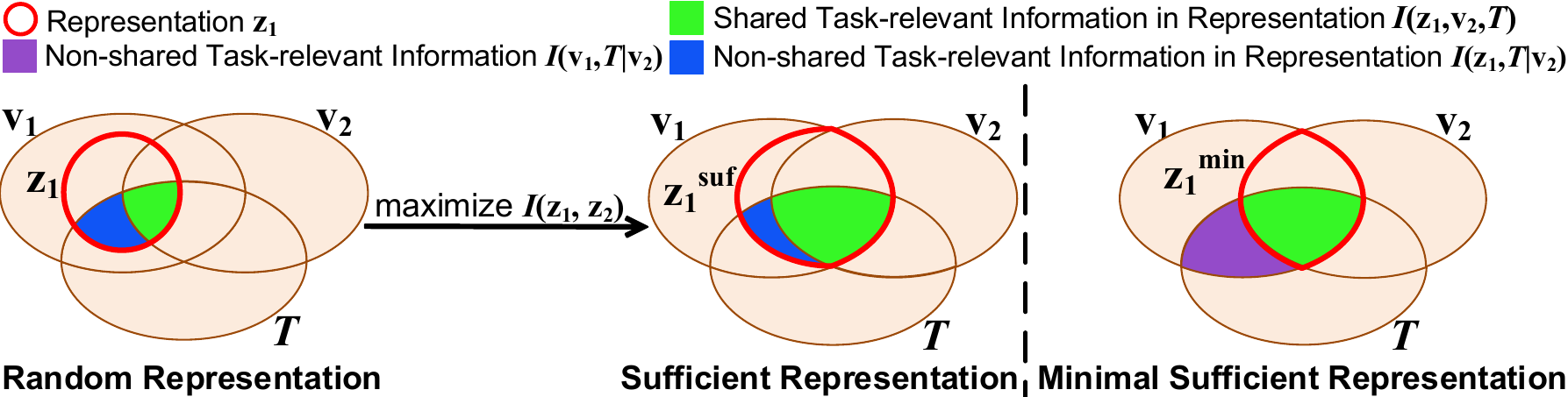}
\caption{Information diagrams of different representations in contrastive learning. We consider the situation where the non-shared task-relevant information $I(v_1,T|v_2)$ cannot be ignored. Contrastive learning makes the representations extracting the shared information between views to obtain the sufficient representation which is approximately minimal. The minimal sufficient representation contains less task-relevant information from the input than other sufficient representations.}
\label{therom_fig}
\end{figure}

The minimal sufficient representation intuitively is not a good choice for downstream tasks, because it completely eliminates the non-shared information between views which may be important for some downstream tasks. We formalize this problem and theoretically prove that in contrastive learning, the minimal sufficient representation is expected to perform worse than other sufficient representations in the downstream tasks. All proofs for the below theorems are provided in Appendix \ref{proofs}.

Considering the symmetry between $v_1$ and $v_2$, without loss of generality, we take $v_2$ as the supervision signal for $v_1$ and take $v_1$ as the input of a task. It is generally believed that the more task-relevant information contained in the representations, the better performance can be obtained \cite{feder1994relations,DBLP:books/daglib/0016881}. Therefore, we examine the task-relevant information contained in the representations.
\begin{theorem}\label{the1}
(Task-Relevant Information in Representations) In contrastive learning, for a downstream task $T$, the minimal sufficient representation $z_1^{min}$ contains less task-relevant information from input $v_1$ than other sufficient representation $z_1^{suf}$, and $I(z_1^{min},T)$ has a gap of $I(v_1,T|v_2)$ with the upper bound $I(v_1,T)$. Formally, we have
\begin{align}
     & I(v_1,T)=I(z_1^{min},T)+I(v_1,T|v_2) \nonumber\\
\geq & I(z_1^{suf},T)=I(z_1^{min},T)+I(z_1^{suf},T|v_2) \nonumber\\
\geq & I(z_1^{min},T)
\end{align}
\end{theorem}
\cref{the1} indicates that $z_1^{suf}$ can have better performance in task $T$ than $z_1^{min}$ because it contains more task-relevant information. When non-shared task-relevant information $I(v_1,T|v_2)$ is significant, $z_1^{min}$ has poor performance because it loses a lot of useful information. See \cref{therom_fig} for the demonstration using information diagrams. To make this observation more concrete, we examine two types of the downstream task: classification tasks and regression tasks, and provide theoretical analysis on the generalization error of the representations.

When the downstream task is a classification task and $T$ is a categorical variable, we consider the Bayes error rate \cite{fukunaga2013introduction} which is the lowest achievable error for any classifier learned from the representations. Concretely, let $P_e$ be the Bayes error rate of arbitrary learned representation $z_1$ and $\widehat{T}$ be the prediction for $T$ based on $z_1$, we have $P_e=1-\mathbb{E}_{p(z_1)}[\max_{t\in T} p(\widehat{T}=t|z_1)]$ and $0\leq P_e\leq 1-1/|T|$ where $|T|$ is the cardinality of $T$. According to the value range of $P_e$, we define a threshold function $\Gamma(x)=\min\{\max\{x,0\},1-1/|T|\}$ to prevent overflow.
\begin{theorem}\label{the2}
(Bayes Error Rate of Representations) For arbitrary learned representation $z_1$, its Bayes error rate $P_e=\Gamma(\Bar{P}_e)$ with
\begin{equation}
\Bar{P}_e\leq 1-\exp[-(H(T)-I(z_1,T|v_2)-I(z_1,v_2,T))]
\end{equation}
Specifically, for sufficient representation $z_1^{suf}$, its Bayes error rate $P_e^{suf}=\Gamma(\Bar{P}_e^{suf})$ with
\begin{equation}\label{suf_bayes}
\resizebox{.9\linewidth}{!}{$\Bar{P}_e^{suf}\leq1-\exp[-(H(T)-I(z_1^{suf},T|v_2)-I(v_1,v_2,T))]$}
\end{equation}
for minimal sufficient representation $z_1^{min}$, its Bayes error rate $P_e^{min}=\Gamma(\Bar{P}_e^{min})$ with
\begin{equation}
\Bar{P}_e^{min}\leq 1-\exp[-(H(T)-I(v_1,v_2,T))]
\end{equation}
\end{theorem}
Since $I(z_1^{suf},T|v_2)\geq0$, \cref{the2} indicates for classification task $T$, the upper bound of $P_e^{min}$ is larger than $P_e^{suf}$. In other words, $z_1^{min}$ is expected to obtain a higher classification error rate in the task $T$ than $z_1^{suf}$. According to the \cref{suf_bayes}, considering that $H(T)$ and $I(v_1,v_2,T)$ are not related to the representations, increasing $I(z_1^{suf},T|v_2)$ can reduce the Bayes error rate in classification task $T$. When $I(z_1^{suf},T|v_2)=I(v_1,T|v_2)$, $z_1^{suf}$ contains all the useful information for task $T$ in $v_1$.

When the downstream task is a regression task and $T$ is a continuous variable, let $\widetilde{T}$ be the prediction for $T$ based on arbitrary learned representation $z_1$, we consider the smallest achievable expected squared prediction error $R_e=\min_{\widetilde{T}}\mathbb{E}[(T-\widetilde{T}(z_1))^2]=\mathbb{E}[\varepsilon^2]$ with $\varepsilon(T,z_1)=T-\mathbb{E}[T|z_1]$.
\begin{theorem}\label{the3}
(Minimum Expected Squared Prediction Error of Representations) For arbitrary learned representation $z_1$, when the conditional distribution $p(\varepsilon|z_1)$ is uniform, Laplacian or Gaussian distribution, the minimum expected squared prediction error $R_e$ satisfies
\begin{equation}
R_e=\alpha\cdot\exp[2\cdot(H(T)-I(z_1,T|v_2)-I(z_1,v_2,T))]
\end{equation}
Specifically, for sufficient representation $z_1^{suf}$, its minimum expected squared prediction error $R_e^{suf}$ satisfies
\begin{equation}
\resizebox{.9\linewidth}{!}{$R_e^{suf}=\alpha\cdot\exp[2\cdot(H(T)-I(z_1^{suf},T|v_2)-I(v_1,v_2,T))]$}
\end{equation}
for minimal sufficient representation $z_1^{min}$, its minimum expected squared prediction error $R_e^{min}$ satisfies
\begin{equation}
R_e^{min}=\alpha\cdot\exp[2\cdot(H(T)-I(v_1,v_2,T))]
\end{equation}
where the constant coefficient $\alpha$ depends on the conditional distribution $p(\varepsilon|z_1)$.
\end{theorem}
The assumption about the estimation error $\varepsilon$ in \cref{the3} is reasonable because $\varepsilon$ is analogous to the `noise' with the mean of 0, which is generally assumed to come from simple distributions (\eg, Gaussian distribution) in statistical learning theory. Similar to the classification tasks, \cref{the3} indicates that for regression tasks, $z_1^{suf}$ can achieve lower expected squared prediction error than $z_1^{min}$ and increasing $I(z_1^{suf},T|v_2)$ can improve the performance.

\cref{the2} and \cref{the3} analyze the disadvantages of the minimal sufficient representation $z_1^{min}$ in classification tasks and regression tasks respectively. The essential reason is that $z_1^{min}$ has less task-relevant information than $z_1^{suf}$ and has a non-ignorable gap $I(v_1,T|v_2)$ with the optimal representation, as shown in \cref{the1}.

\subsection{More non-shared task-relevant information}
According to the above theoretical analysis, in contrastive learning, the minimal sufficient representation is not sufficient for downstream tasks due to the lack of some non-shared task-relevant information. Moreover, contrastive learning approximately learns the minimal sufficient representation, thereby having the risk of over-fitting to the shared information between views. To this end, we propose to extract more non-shared task-relevant information from $v_1$, \ie, increasing $I(z_1,T|v_2)$. However, we cannot utilize any downstream task information during training, so it is impossible to increase $I(z_1,T|v_2)$ directly. We consider increasing $I(z_1,v_1)$ as an alternative because the increased information from $v_1$ in $z_1$ may be relevant to some downstream tasks, and this motivation is demonstrated in \cref{motivation}. In addition, increasing $I(z_1,v_1)$ also helps to extract the shared information between views at the beginning of the optimization process. Concretely, considering the symmetry between $v_1$ and $v_2$, our optimization objective is
\begin{equation}
    \max_{f_1,f_2} I(z_1,z_2)+\sum_{i=1}^2\lambda_i I(z_i,v_i)
\end{equation}
which consists of the original optimization objective \cref{orig} in contrastive learning and the objective terms we proposed. The coefficients $\lambda_1$ and $\lambda_2$ are used to control the amount of increasing $I(z_1,v_1)$ and $I(z_2,v_2)$ respectively. For optimizing $I(z_1,z_2)$, we adopt the commonly used implementations in contrastive learning models \cite{chen2020simple,DBLP:conf/nips/GrillSATRBDPGAP20,DBLP:conf/icml/ZbontarJMLD21}. For optimizing $I(z_i,v_i),i=1,2$, we consider two implementations.

\paragraph{Implementation \uppercase\expandafter{\romannumeral1}}
Since $I(z,v)=H(v)-H(v|z)$ and $H(v)$ is not related with $z$, we can equivalently decrease the conditional entropy $H(v|z)=-\mathbb{E}_{p(z,v)}[\ln p(v|z)]$. Concretely, we use the representation $z$ to reconstruct the original input $v$, as done in auto-encoder models \cite{DBLP:journals/jmlr/VincentLLBM10}. Decreasing the entropy of reconstruction encourages the representation $z$ to contain more information about the original input $v$. However, the conditional distribution $p(v|z)$ is intractable in practice, so we use $q(v|z)$ as an approximation and get $\mathbb{E}_{p(z,v)}[\ln q(v|z)]$, which is the lower bound of $\mathbb{E}_{p(z,v)}[\ln p(v|z)]$. We can increase $\mathbb{E}_{p(z,v)}[\ln q(v|z)]$ as an alternative objective. According to the type of input $v$ (\eg, images, text or audio), $q(v|z)$ can be any appropriate distribution with known probability density function, such as Bernoulli distribution, Gaussian distribution or Laplace distribution, and its parameters are the functions of $z$. For example, when $q(v|z)$ is the Gaussian distribution $\mathcal{N}(v;\mu(z),\sigma^2 I)$ with given variance $\sigma^2$ and deterministic mean function $\mu(\cdot)$ which is usually parameterized by neural networks, we have
\begin{equation}
    \mathbb{E}_{p(z,v)}[\ln q(v|z)]\propto -\mathbb{E}_{p(z,v)}[\|v-\mu(z)\|_2^2]+c
\end{equation}
where $c$ is a constant to representation $z$. The final optimization objective is
\begin{equation}\label{imp1}
\max_{f_1,f_2,\mu} I(z_1,z_2)-\sum_{i=1}^2\lambda_i\mathbb{E}_{p(z_i,v_i)}[\|v_i-\mu_i(z_i)\|_2^2]
\end{equation}

\paragraph{Implementation \uppercase\expandafter{\romannumeral2}}
Although the above implementation is effective and preferred in practice, it needs to reconstruct the input, which is challenging for complex input and introduces more model parameters. To this end, we propose another representation-level implementation as an optional alternative. We investigate various lower bound estimates of mutual information, such as the bound of Barber and Agakov \cite{barber2003algorithm}, the bound of Nguyen, Wainwright and Jordan \cite{nguyen2010estimating}, MINE \cite{belghazi2018mutual} and InfoNCE \cite{poole2019variational}. We choose the InfoNCE lower bound and the detailed discussion is provided in Appendix \ref{Choice_LBE}. Concretely, the InfoNCE lower bound is
\begin{equation}
    \Hat{I}_{NCE}(z,v)=\mathbb{E}\left[\frac{1}{N}\sum_{k=1}^N\ln\frac{p(z^k|v^k)}{\frac{1}{N}\sum_{l=1}^Np(z^l|v^k)}\right]
\end{equation}
where $(z^k,v^k),k=1,\cdots,N$ are $N$ copies of $(z,v)$ and the expectation is over $\Pi_kp(z^k,v^k)$. In the implementation \uppercase\expandafter{\romannumeral1}, we map the input $v$ to the representation $z$ through a deterministic function $f$ with $z=f(v)$. Differently, here we need the expression of $p(z|v)$ to calculate the InfoNCE lower bound, which means the representation $z$ is no longer a deterministic output of input $v$, so we use the reparameterization trick \cite{DBLP:journals/corr/KingmaW13} during training. For example, when we define $p(z|v)$ as the Gaussian distribution $\mathcal{N}(z;f(v),\sigma^2 I)$ with given variance $\sigma^2$ and the function $f$ is the same as in the Implementation \uppercase\expandafter{\romannumeral1}, we have $z=f(v)+\epsilon\sigma,\epsilon\sim\mathcal{N}(\bm{0},I)$ and $\Hat{I}_{NCE}$ is equivalent to
\begin{equation}\label{reg}
\resizebox{.88\linewidth}{!}{$\tilde{I}_{NCE}(z,v)=\mathbb{E}\left[-\frac{1}{N}\sum_{k=1}^N\ln\sum_{l=1}^N\exp(-\rho\|z^l-f(v^k)\|_2^2)\right]$}
\end{equation}
where $\rho$ is a scale factor. In fact, it pushes the representations away from each other to increase $H(z)$, which can increase mutual information $I(z,v)$ since $I(z,v)=H(z)-H(z|v)=H(z)-\frac{d}{2}(\ln{2\pi}+\ln{\sigma^2}+1)$ with $d$ being representation dimension. It also be denoted as uniformity property \cite{wang2020understanding}. The final optimization objective is
\begin{equation}\label{imp2}
    \max_{f_1,f_2} I(z_1,z_2)+\sum_{i=1}^2\lambda_i\tilde{I}_{NCE}(z_i,v_i)
\end{equation}
Since the objective term \cref{reg} is calculated at the representation-level, when we use the convolutional neural networks (\eg, ResNet \cite{he2016deep}) to parameterize $f$, it can be applied to the output activation of multiple internal blocks.

\paragraph{Discussion.}
It is worth noting that increasing $I(z,v)$ does not conflict with the information bottleneck theory \cite{tishby2015deep}. According to our analysis, the learned representations in contrastive learning are not sufficient for the downstream tasks. Therefore, we need to make the information in the representations more sufficient but not to compress it. On the other hand, we cannot introduce too much information from the input $v$ either, which may contain harmful noise. Here we use the coefficients $\lambda_1$ and $\lambda_2$ to control this.

\section{Experiments}
In this section, we first verify the effectiveness of increasing $I(z,v)$ on various datasets, and then provide some analytical experiments. We choose three classic contrastive learning models as our baselines: SimCLR \cite{chen2020simple}, BYOL \cite{grill2020bootstrap} and Barlow Twins \cite{DBLP:conf/icml/ZbontarJMLD21}. We denote our first implementation \cref{imp1} as "RC" for "ReConstruction" and the second implementation \cref{imp2} as "LBE" for "Lower Bound Estimate". For all experiments, we use random cropping, flip and random color distortion as the data augmentation, as suggested by \cite{chen2020simple}. For "LBE", we set $\sigma=0.1$ and $\rho=0.05$.

\subsection{Effectiveness of increasing $I(z,v)$}
We consider different types of the downstream task, including classification, detection and segmentation tasks. The results of Barlow Twins are provided in Appendix \ref{bartwins}.

\begin{table*}
\centering
\scalebox{0.9}{
\begin{tabular}{l|c|cccccc}
\toprule[0.1em]
Model      & CIFAR10 & DTD & MNIST & FaMNIST & CUBirds & VGGFlower & TrafficSigns \\ \hline
SimCLR     & 85.76 & 29.52 & 97.03 & 88.36 & 8.87  & 42.81 & 92.41 \\
SimCLR+RC (ours)  & \textbf{85.78} & 33.67 & \textbf{97.99} & \textbf{90.31} & \textbf{10.89} & \textbf{54.16} & \textbf{95.84} \\
SimCLR+LBE (ours) & 85.45 & \textbf{34.52} & 97.94 & 89.26 & 10.60 & 54.10 & 94.96 \\ \hline
BYOL       & 85.64 & 31.22 & 97.15 & 88.92 & 8.84  & 40.90 & 92.17 \\
BYOL+RC (ours)  & \textbf{85.80} & \textbf{34.73} & \textbf{98.07} & \textbf{89.61} & 9.68  & 48.75 & 94.19 \\
BYOL+LBE (ours)   & 85.28 & 33.99 & 97.76 & 88.99 & \textbf{9.96} & \textbf{54.10} & \textbf{95.09} \\
\toprule[0.1em]
Model      & STL-10 & DTD & MNIST & FaMNIST & CUBirds & VGGFlower & TrafficSigns \\ \hline
SimCLR     & 78.74  & 39.41 & 95.00 & 87.31 & 8.34  & 49.41 & 80.25 \\
SimCLR+RC (ours)  & 79.21  & 41.81 & \textbf{97.48} & \textbf{89.98} & 10.03 & \textbf{60.46} & \textbf{94.73} \\
SimCLR+LBE (ours) & \textbf{80.17}  & \textbf{42.07} & 97.04 & 88.68 & \textbf{10.11} & 58.51 & 87.77 \\ \hline
BYOL       & 80.83  & 40.05 & 94.45 & 87.23 & 8.54  & 49.41 & 77.54 \\
BYOL+RC (ours)    & \textbf{81.11}  & 42.02 & \textbf{96.96} & \textbf{88.92} & 9.63  & 55.71 & \textbf{88.57} \\
BYOL+LBE (ours)   & 80.85 & \textbf{42.55} & 95.75 & 87.88 & \textbf{10.55} & \textbf{59.39} & 84.62 \\
\toprule[0.1em]
Model      & ImageNet & DTD & CIFAR10 & CIFAR100 & CUBirds & VGGFlower & TrafficSigns \\ \hline
SimCLR     & 61.01 & 70.16 & 82.30 & 59.86 & 36.49 & 93.52 & 95.27 \\
SimCLR+RC (ours)  & \textbf{61.60} & \textbf{71.22} & \textbf{83.30} & \textbf{63.56} & 37.42 & \textbf{94.53} & \textbf{96.47} \\
SimCLR+LBE (ours) & 61.37 & 70.95 & 83.20 & 61.99 & \textbf{37.78} & 94.34 & 95.99 \\
\hline
\end{tabular}}
\caption{Linear evaluation accuracy ($\%$) on the source dataset (CIFAR10, STL-10 or ImageNet) and other transfer datasets.}
\label{LE_result}
\vspace{-2mm}
\end{table*}

\paragraph{Pretraining.}
We train the models on CIFAR10 \cite{krizhevsky2009learning}, STL-10 \cite{coates2011analysis} and ImageNet \cite{deng2009imagenet}. For CIFAR10 and STL-10, we use the ResNet18 \cite{he2016deep} backbone and the models are trained for 200 epochs with batch size 256 using Adam optimizer with learning rate 3e-4. For ImageNet, we use the ResNet50 \cite{he2016deep} backbone and the models are trained for 200 epochs with batch size 1024 using LARS optimizer \cite{you2017scaling} and a cosine decay learning rate schedule.
% We set $\lambda_1=\lambda_2=1$ for SimCLR and $\lambda_1=\lambda_2=0.1$ for BYOL.

\paragraph{Linear evaluation.}
We follow the linear evaluation protocol where a linear classifier is trained on top of the frozen backbone. The linear evaluation is conducted on the source dataset and several transfer datasets: CIFAR100 \cite{krizhevsky2009learning}, DTD \cite{cimpoi2014describing}, MNIST \cite{lecun1998gradient}, FashionMNIST \cite{xiao2017fashion}, CUBirds \cite{wah2011caltech}, VGG Flower \cite{nilsback2008automated} and Traffic Signs \cite{houben2013detection}. The linear classifier is trained for 100 epochs using SGD optimizer. \cref{LE_result} shows the results on CIFAR10, STL-10 and ImageNet, and the best result in each block is in bold. Our implemented results of the baselines are consistent with \cite{wang2021understanding,DBLP:conf/iclr/TamkinWG21,DBLP:conf/nips/HoN20,chen2020simple}. Increasing $I(z,v)$ can introduce non-shared information and improve the classification accuracy, especially on transfer datasets. This means the shared information between views is not sufficient for some tasks, \eg, classification on DTD, VGG Flower and Traffic Signs where increasing $I(z,v)$ achieves significant improvement. In other words, increasing $I(z,v)$ can prevent the models from over-fitting to the shared information between views. What's more, it is effective for various contrastive learning models, which means our analysis results are widely applicable in contrastive learning. In fact, they all satisfy the internal mechanism.

\begin{table}
\footnotesize
\centering
\subfloat[Object detection on VOC07+12]{\label{voc}
\begin{tabularx}{0.45\textwidth}{l|c|c|c}
Model & AP & AP$_{50}$ & AP$_{75}$ \\
\toprule[0.1em]
random initialization & 33.8 & 60.2 & 33.1 \\
\hline
SimCLR             & 45.5 & 76.2 & 47.5 \\
SimCLR+RC (ours)   & 48.1 {\color{magenta}(+2.6)} & 78.0 {\color{magenta}(+1.8)} & 50.9 {\color{magenta}(+3.4)} \\
SimCLR+LBE (ours)  & 47.4 {\color{magenta}(+1.9)} & 77.8 {\color{magenta}(+1.6)} & 49.7 {\color{magenta}(+2.2)} \\
\end{tabularx}}

\subfloat[Object detection on COCO]{\label{coco_det}
\begin{tabularx}{0.45\textwidth}{l|c|c|c}
Model & AP$^{bb}$ & AP$_{50}^{bb}$ & AP$_{75}^{bb}$ \\
\toprule[0.1em]
random initialization & 35.6 & 54.6 & 38.2 \\
\hline
SimCLR             & 38.6 & 58.5 & 41.8 \\
SimCLR+RC (ours)   & 39.3 {\color{magenta}(+0.7)} & 59.0 {\color{magenta}(+0.5)} & 42.6 {\color{magenta}(+0.8)} \\
SimCLR+LBE (ours)  & 39.0 (+0.4) & 58.7 (+0.2) & 42.3 {\color{magenta}(+0.5)} \\
\end{tabularx}}

\subfloat[Instance segmentation on COCO]{\label{coco_seg}
\begin{tabularx}{0.45\textwidth}{l|c|c|c}
Model & AP$^{mk}$ & AP$_{50}^{mk}$ & AP$_{75}^{mk}$ \\
\toprule[0.1em]
random initialization & 31.4 & 51.5 & 33.5 \\
\hline
SimCLR             & 33.9 & 55.2 & 36.0 \\
SimCLR+RC (ours)   & 34.5 {\color{magenta}(+0.6)} & 55.7 {\color{magenta}(+0.5)} & 36.6 {\color{magenta}(+0.6)} \\
SimCLR+LBE (ours)  & 34.2 (+0.3) & 55.4 (+0.2) & 36.4 (+0.4) \\
\end{tabularx}}
\caption{Object detection and instance segmentation on VOC07+12 and COCO. The models on COCO are fine-tuned using the default $2\times$ schedule. In magenta are the gaps of at least {\color{magenta}+0.5} point to the baseline, SimCLR.}
\label{obj_det}
\end{table}

\paragraph{Object detection and instance segmentation.}
We conduct object detection on VOC07+12 \cite{everingham2010pascal} using Faster R-CNN \cite{ren2015faster}, and detection and instance segmentation on COCO \cite{lin2014microsoft} using Mask R-CNN \cite{he2017mask}, following the setup in \cite{he2020momentum}. All methods use the R50-C4 \cite{he2017mask} backbone that is initialized using the ResNet50 pre-trained on ImageNet. The results are show in \cref{obj_det}. Increasing $I(z,v)$ significantly improves the precision in object detection and instance segmentation tasks. These dense prediction tasks require some local semantic information from the input. Increasing $I(z,v)$ can make the representation $z$ contain more information from the input $v$ which may not be shared between views, thereby obtaining better precision.

\begin{table*}[ht]
\centering
\scalebox{0.9}{
\begin{tabular}{l|c|cccccc}
\toprule[0.1em]
Model      & CIFAR10 & DTD & MNIST & FaMNIST & CUBirds & VGGFlower & TrafficSigns \\
\midrule[0.1em]
SimCLR     & 85.76 & 29.52 & \textbf{97.03} & \textbf{88.36} & \textbf{8.87} & 42.81 & \textbf{92.41} \\
SimCLR+IP  & \textbf{85.86} & \textbf{30.15} & 96.71 & 88.18 & 8.66 & \textbf{43.22} & 92.13 \\
\hline
SimCLR$^{\dag}$ & 85.81 & \textbf{31.70} & \textbf{97.08} & \textbf{88.85} & 8.77 & \textbf{44.41} & 92.41 \\
SimCLR+MIB  & \textbf{86.20} & 31.17 & 97.00 & 88.62 & \textbf{9.01} & 43.88 & \textbf{93.01}\\
\bottomrule[0.1em]
\end{tabular}}
\caption{Linear evaluation accuracy ($\%$) on CIFAR10 and the transfer datasets. $\dag$ represents adding Gaussian noise to the representations.}
\label{mib_ip}
\end{table*}

\subsection{Analytical experiments}\label{exp_ana}
We provide some analytical experiments to further understand our hypotheses, theoretical analysis and models.

\begin{figure}
    \centering
    \subcaptionbox{CIFAR10}{
        \includegraphics[width=0.23\textwidth]{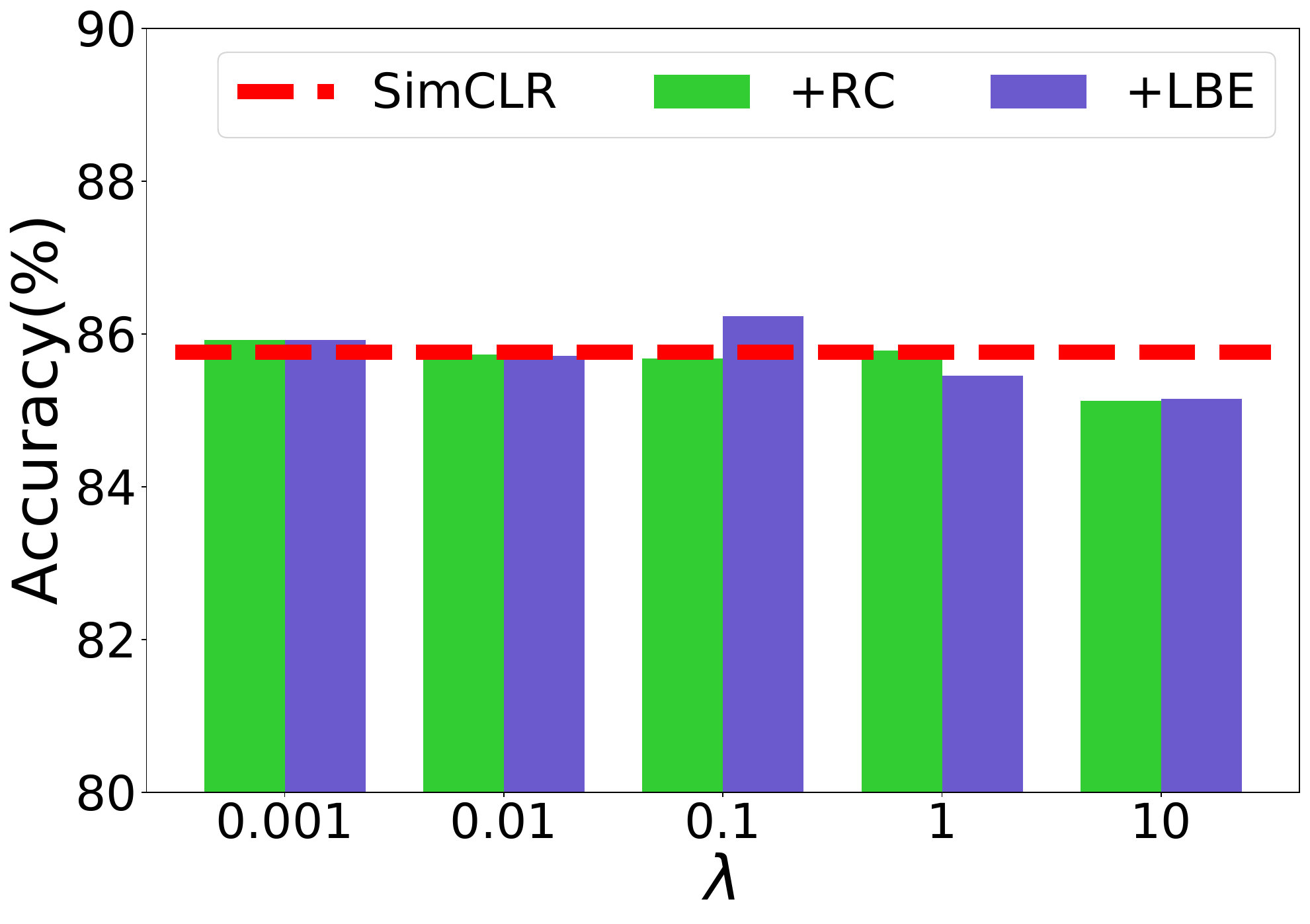}}
    \subcaptionbox{Transfer from CIFAR10}{
        \includegraphics[width=0.23\textwidth]{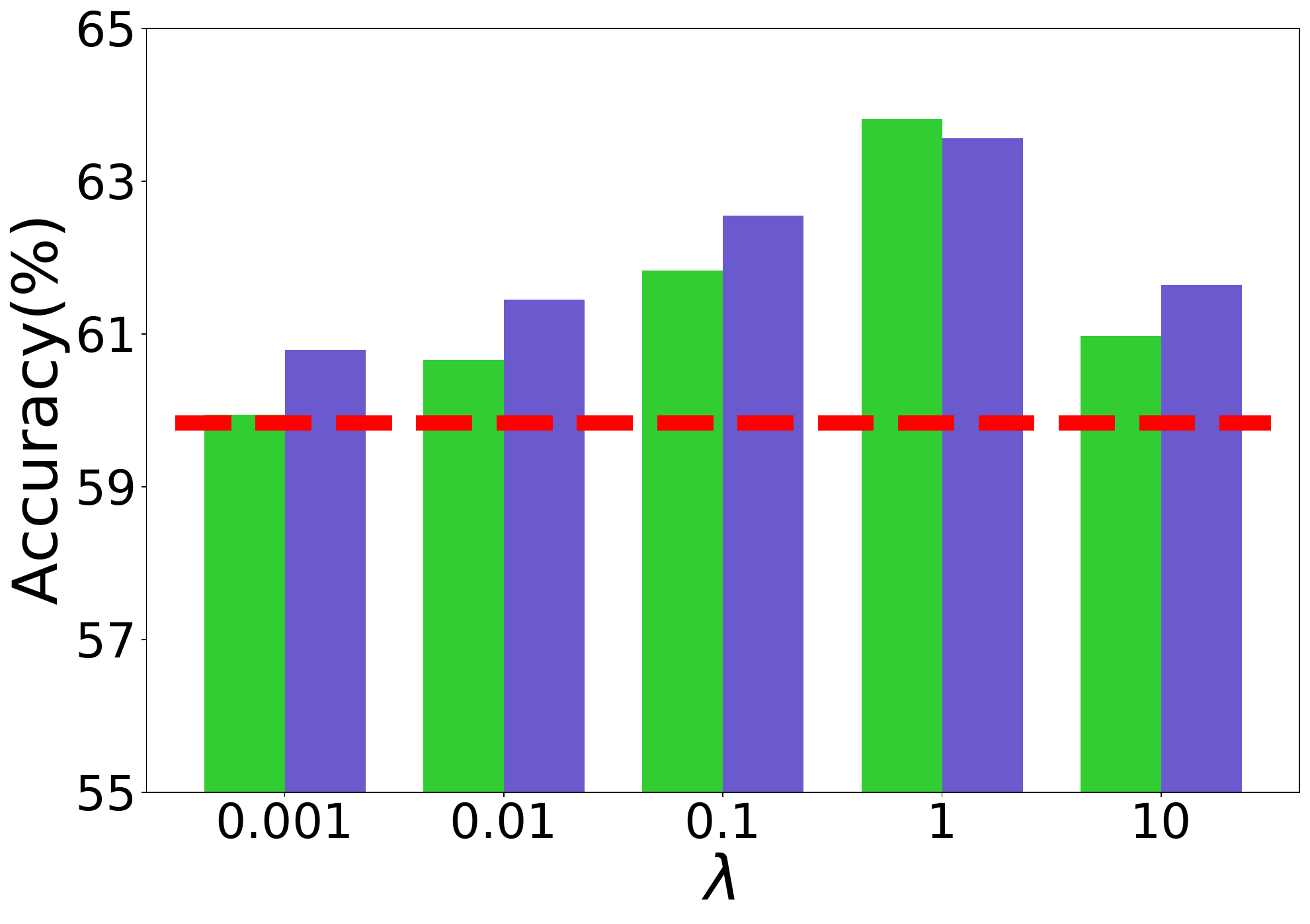}}
    \subcaptionbox{STL-10}{
        \includegraphics[width=0.23\textwidth]{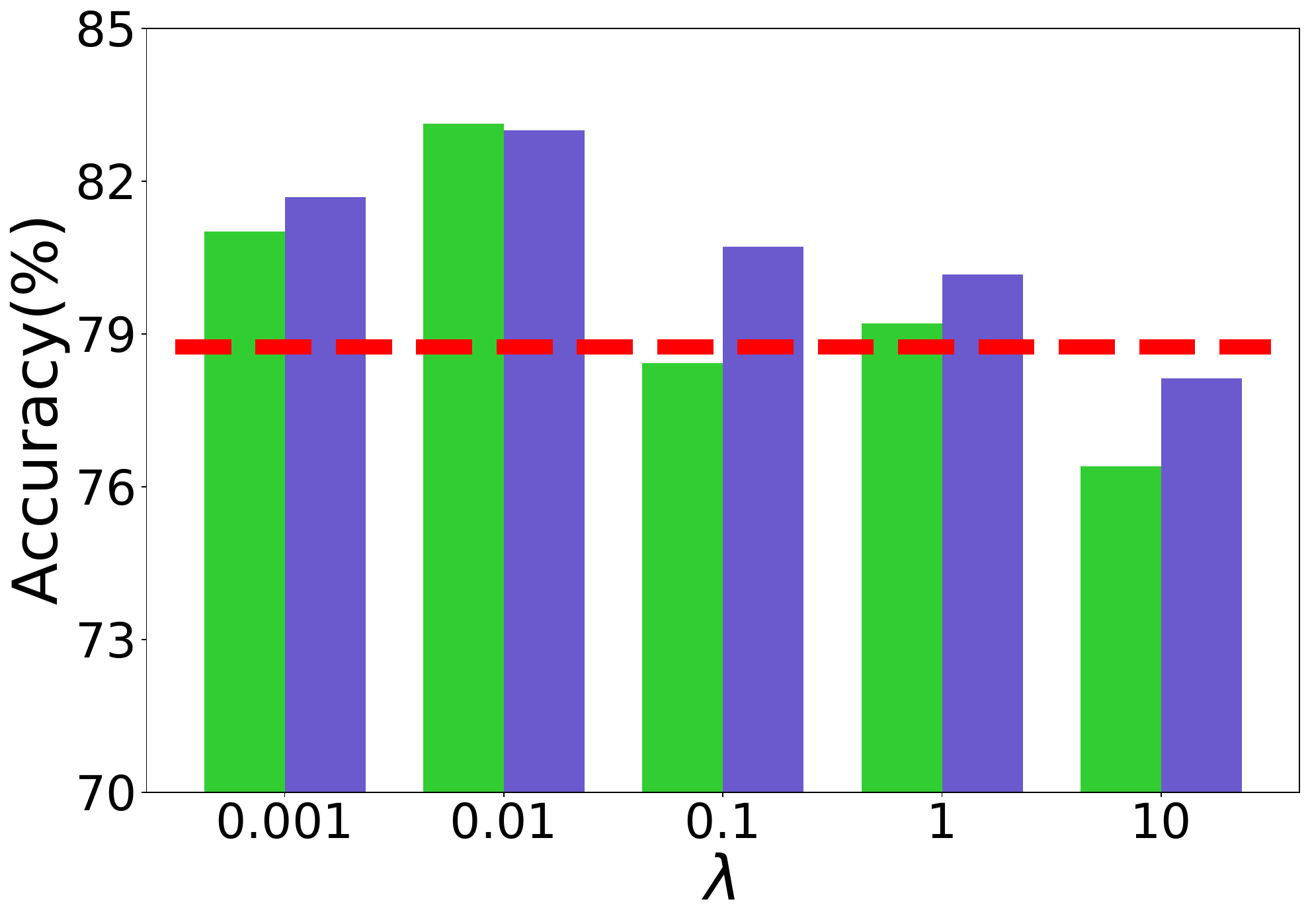}}
    \subcaptionbox{Transfer from STL-10}{
        \includegraphics[width=0.23\textwidth]{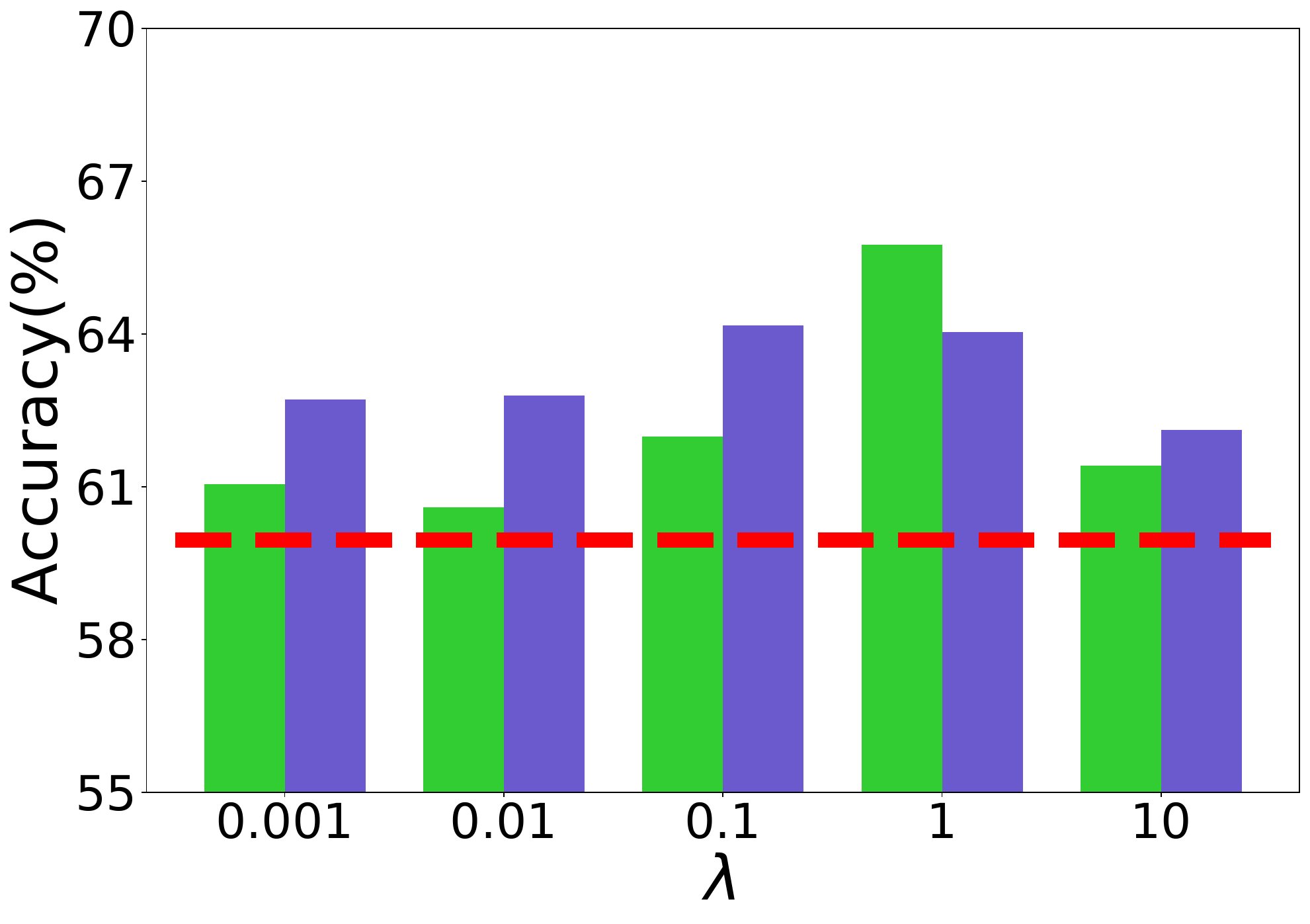}}
    \caption{Linear evaluation accuracy on the source dataset (CIFAR10 or STL-10) and the averaged accuracy on all transfer datasets with varying hyper-parameter $\lambda$.}
    \label{lambda}
\end{figure}

\paragraph{Eliminating non-shared information.}
Some recent works \cite{DBLP:conf/iclr/Federici0FKA20,DBLP:conf/iclr/Tsai0SM21} propose to eliminate the non-shared information between views in the representation to get the minimal sufficient representation. To this end, Federici \etal \cite{DBLP:conf/iclr/Federici0FKA20} minimize the regularization term
\begin{equation}
\resizebox{.88\linewidth}{!}{$L_{MIB}=\frac{1}{2}\left[KL(p(z_1|v_1)\Vert p(z_2|v_2))+KL(p(z_2|v_2)\Vert p(z_1|v_1))\right]$}
\end{equation}
where $KL(\cdot||\cdot)$ represents the Kullback-Leibler divergence. When $p(z_1|v_1)$ and $p(z_2|v_2)$ are modeled as $\mathcal{N}(z_i;f_i(v_i),\sigma^2I),i=1,2$ with given variance $\sigma^2$, it can be rewritten as $L_{MIB}=\mathbb{E}_{p(v_1,v_2)}\left[\|f_1(v_1)-f_2(v_2)\|_2^2\right]$. Identically, Tsai \etal \cite{DBLP:conf/iclr/Tsai0SM21} minimize the inverse predictive loss $L_{IP}=\mathbb{E}_{p(v_1,v_2)}\left[\|f_1(v_1)-f_2(v_2)\|_2^2\right]$. The detailed derivation is provided in Appendix \ref{MIB_IP}. We evaluate these two regularization terms in the linear evaluation tasks and choose their coefficient with best accuracy on the source dataset. The results are shown in \cref{mib_ip} and the best result in each block is in bold. Although these two regularization terms have the same form, $L_{MIB}$ uses stochastic encoders which is equivalent to adding Gaussian noise, so we report the results of SimCLR with Gaussian noise, marked by $\dag$. As we can see, eliminating the non-shared information cannot change the accuracy in downstream classification tasks much. This means that the sufficient representation learned in contrastive learning is approximately minimal and we don’t need to further remove the non-shared information.

\begin{figure}
    \centering
    \subcaptionbox{CIFAR10}{
        \includegraphics[width=0.23\textwidth]{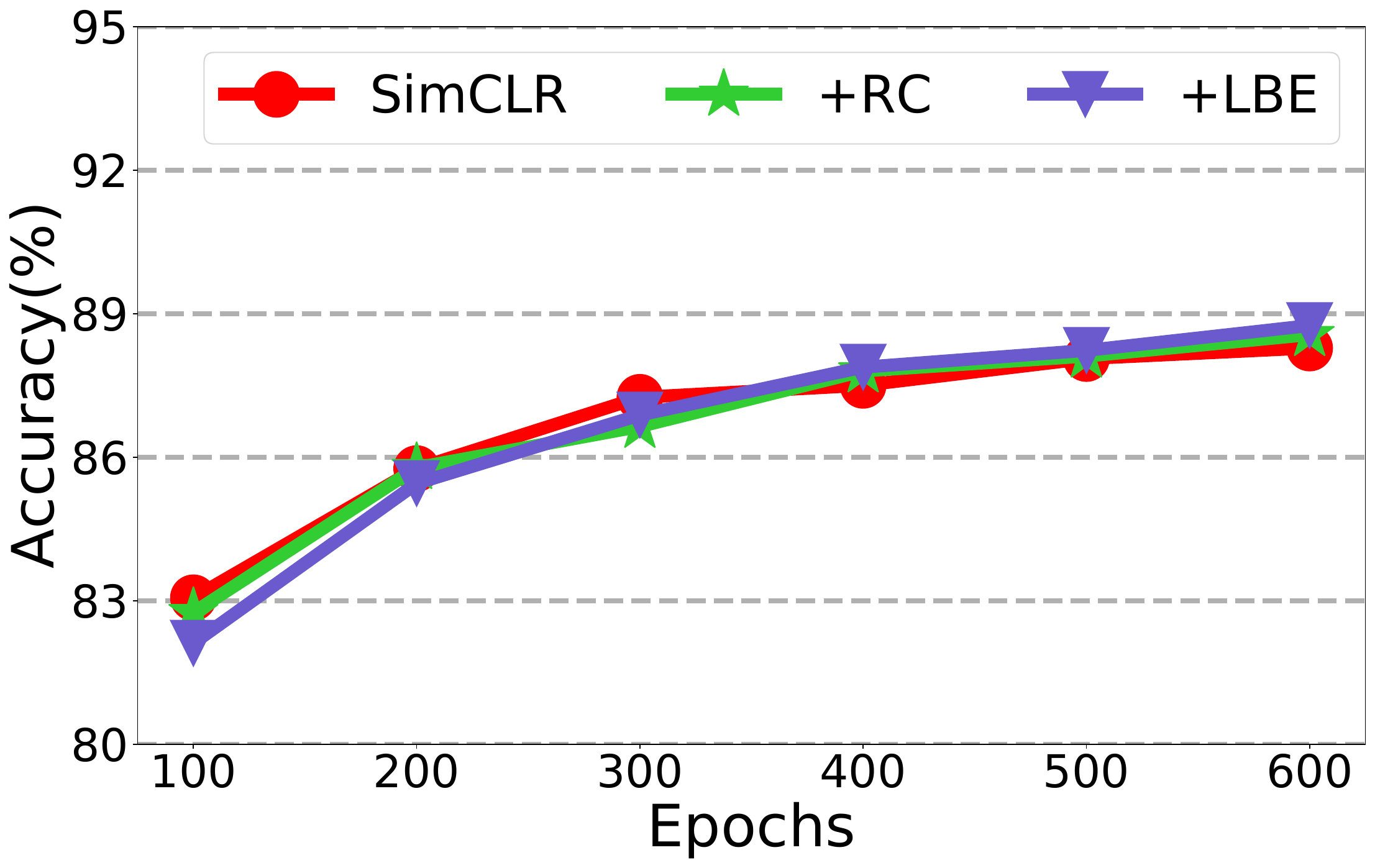}}
    \subcaptionbox{Transfer from CIFAR10}{
        \includegraphics[width=0.23\textwidth]{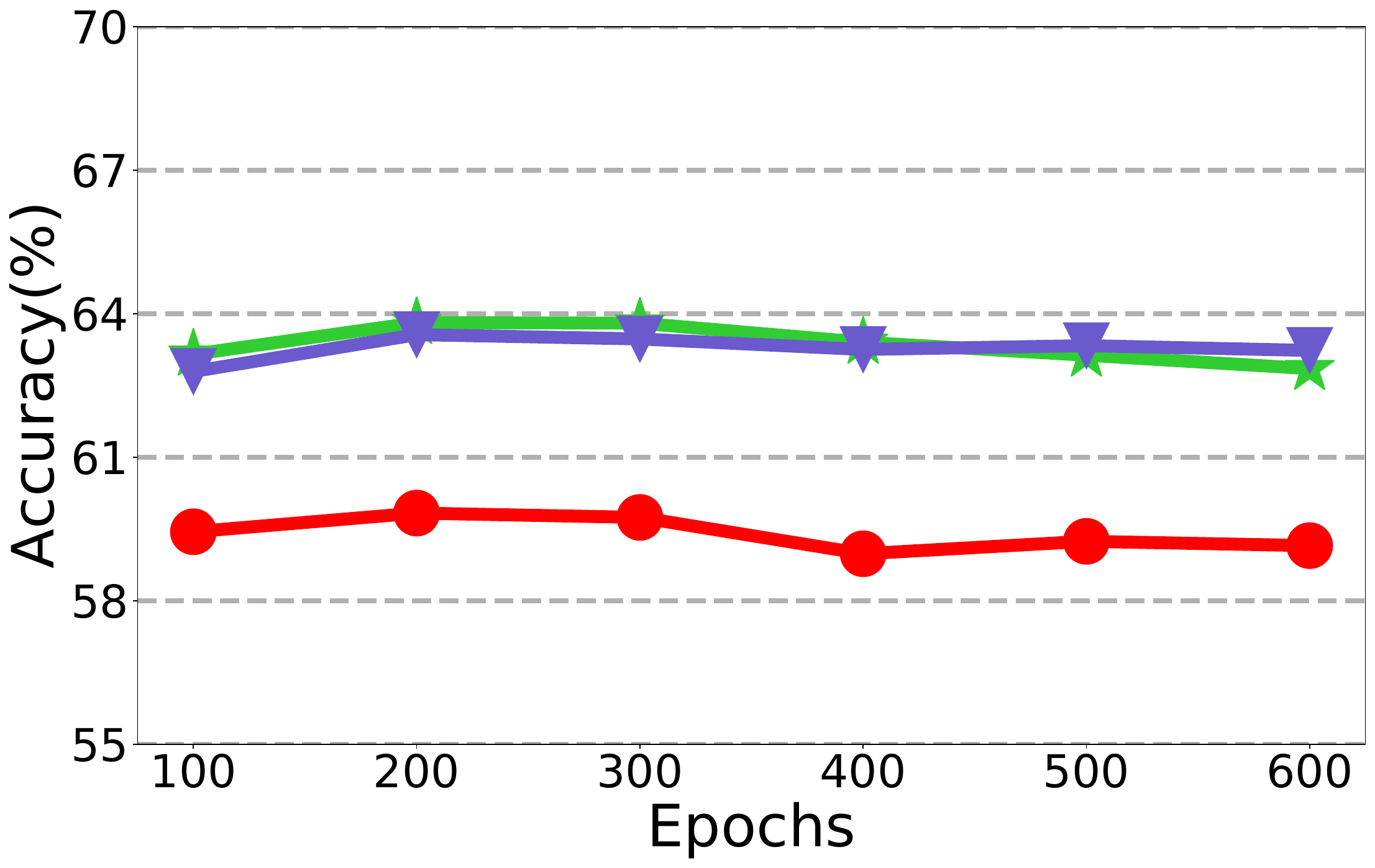}}
    \subcaptionbox{STL-10}{
        \includegraphics[width=0.23\textwidth]{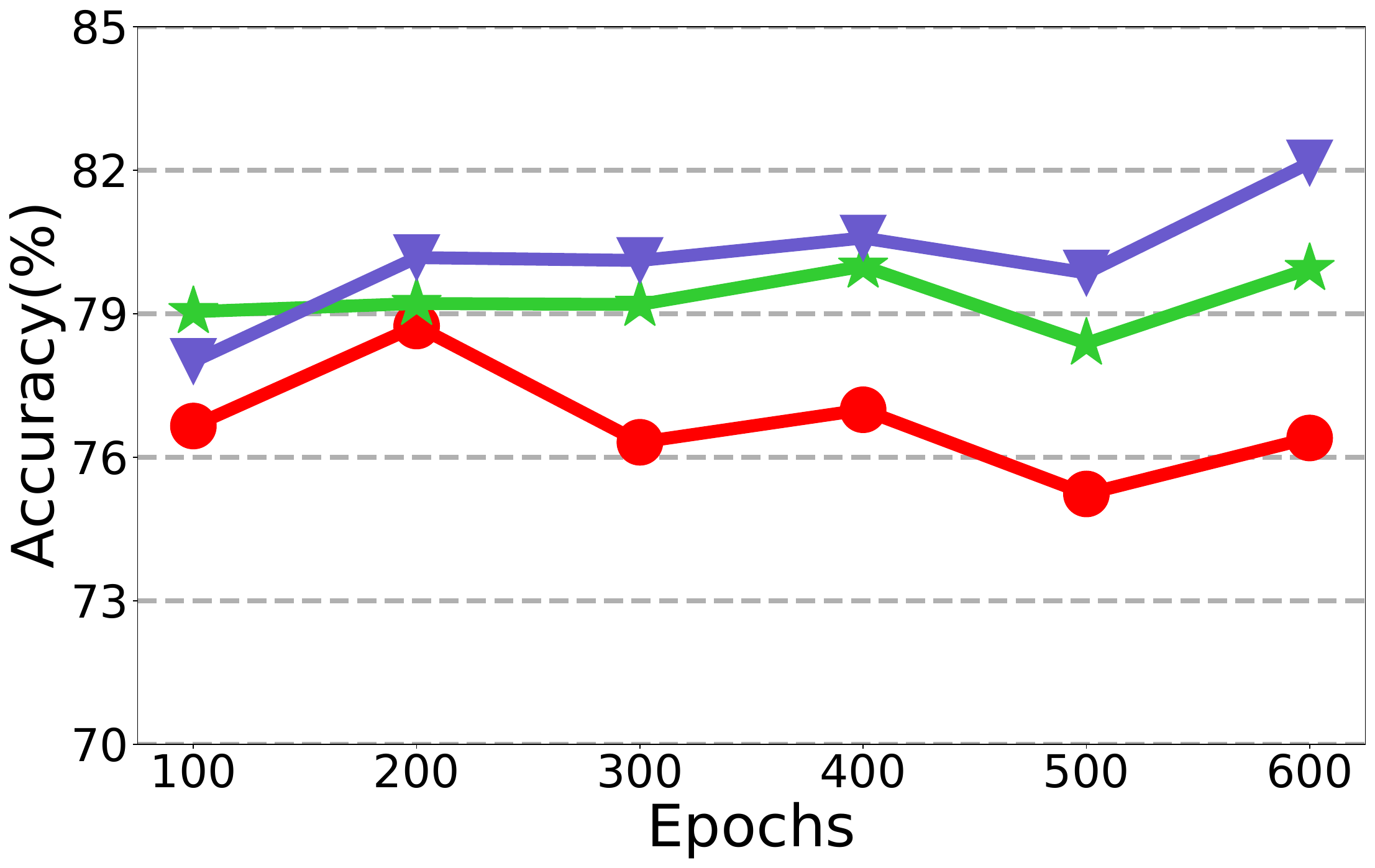}}
    \subcaptionbox{Transfer from STL-10}{
        \includegraphics[width=0.23\textwidth]{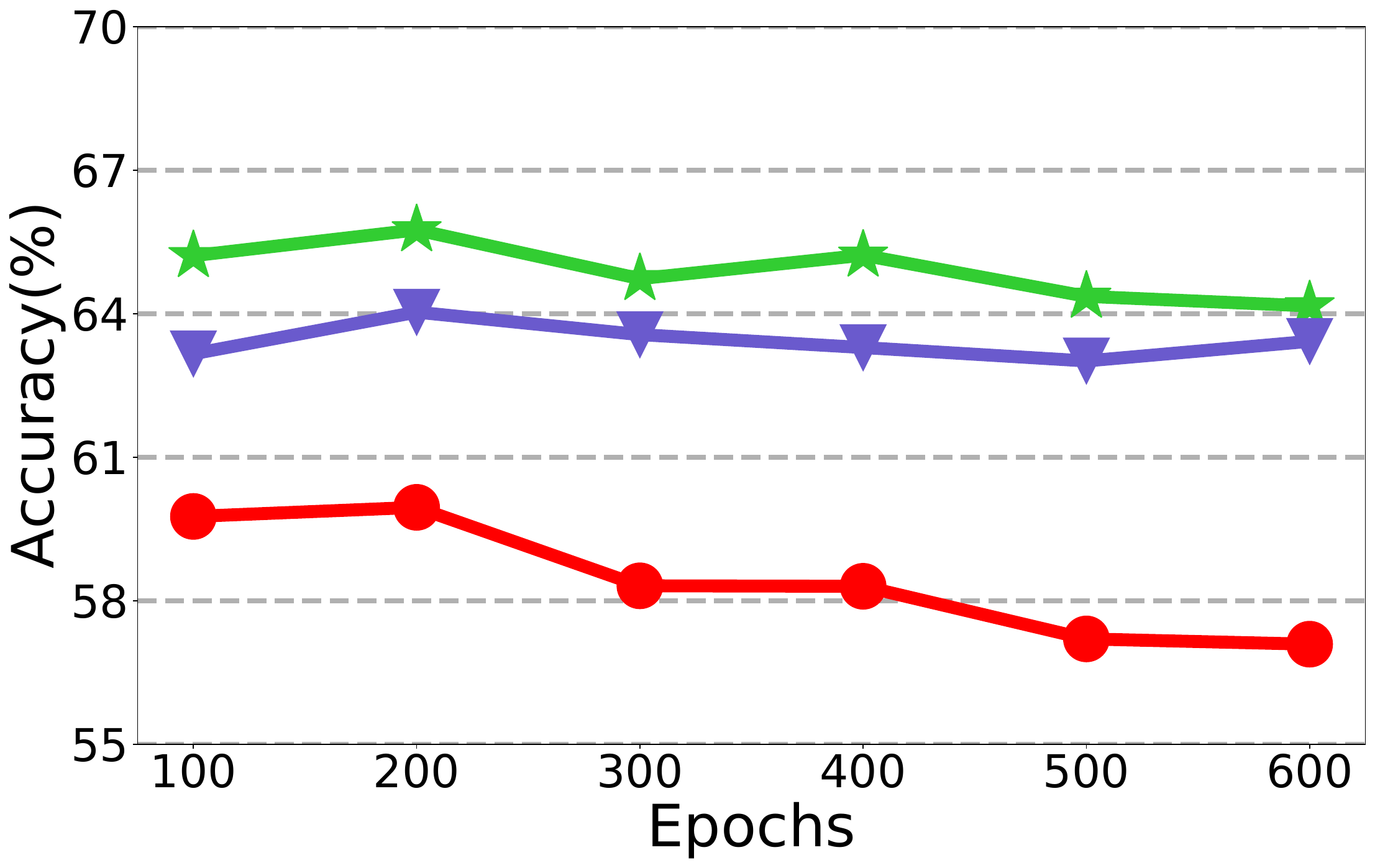}}
    \caption{Linear evaluation accuracy on the source dataset (CIFAR10 or STL-10) and the averaged accuracy on all transfer datasets with varying epochs.}
    \label{epoch}
\end{figure}

\paragraph{Changing the amount of increasing $I(z,v)$.}
Quantifying the mutual information between the high-dimensional variables is very difficult, and often leads to inaccurate calculation in practice \cite{DBLP:conf/iclr/SongE20,mcallester2020formal}. Therefore, we assume that the hyper-parameters $\lambda_1$ and $\lambda_2$ control the amount of increasing $I(z_1,v_1)$ and $I(z_2,v_2)$ respectively. Larger $\lambda_1$ is expected to increase $I(z_1,v_1)$ more, so as $\lambda_2$. We set $\lambda_1=\lambda_2=\lambda$ and evaluate the performance of different $\lambda$ from $\{0.001,0.01,0.1,1,10\}$. We choose SimCLR as the baseline and the results are shown in \cref{lambda}. We report the accuracy on the source dataset (CIFAR10 or STL-10) and the averaged accuracy on all transfer datasets. As we can see, increasing $I(z,v)$ consistently improves the performance in downstream classification tasks. We can observe a non-monotonous reverse-U trend of accuracy with the change of $\lambda$, which means excessively increasing $I(z,v)$ may introduce noise beside useful information.

\begin{table*}
\centering
\scalebox{0.9}{
\begin{tabular}{l|c|cccccc}
\toprule[0.1em]
Model      & CIFAR10 & DTD & MNIST & FaMNIST & CUBirds & VGGFlower & TrafficSigns \\
\hline
Supervised            & \textbf{93.25} & 34.10 & 98.52 & 90.09 & 8.37 & 46.14 & 93.05 \\
Supervised+RC (ours)  & 93.09 & 32.77 & 98.61 & 89.77 & 8.84 & 49.05 & 93.28 \\
Supervised+LBE (ours) & 93.18 & \textbf{34.79} & \textbf{98.68} & \textbf{90.40} & \textbf{9.72} & \textbf{53.15} & \textbf{94.47} \\
\toprule[0.1em]
Model      & CIFAR100 & DTD & MNIST & FaMNIST & CUBirds & VGGFlower & TrafficSigns \\
\hline
Supervised            & 71.92 & 36.06 & 98.48 & 88.97 & 11.51 & 64.21 & 96.54 \\
Supervised+RC (ours)  & \textbf{72.02} & 34.79 & \textbf{98.59} & 89.35 & 10.94 & 65.34 & 96.67 \\
Supervised+LBE (ours) & 71.89 & \textbf{36.33} & 98.37 & \textbf{89.42} & \textbf{11.89} & \textbf{65.64} & \textbf{96.91} \\
\hline
\end{tabular}}
\caption{Linear evaluation accuracy ($\%$) on the source dataset (CIFAR10 or CIFAR100) and the transfer datasets.}
\label{super}
\end{table*}

\paragraph{Training with more epochs.}
In the above experiments, we train all models for 200 epochs. Here we further show the behavior of the contrastive learning models and increasing $I(z,v)$ when training with more epochs. We choose SimCLR as the baseline and train all models for 100, 200, 300, 400, 500 and 600 epochs. The results are shown in \cref{epoch}. With more training epochs, the learned representations in contrastive learning are more approximate to the minimal sufficient representation which mainly contain the shared information between views and ignore the non-shared information. For the classification tasks on the transfer datasets, the shared information between views is not sufficient. As shown in \cref{epoch} (b) and (d), the accuracy on the transfer datasets decreases with more epochs and the learned representations over-fit to the shared information between views. Increasing $I(z,v)$ can introduce non-shared information and obtain the significant improvement. For the classification tasks on the source datasets, the shared information between views is sufficient on CIFAR10 but not on STL-10. As shown in \cref{epoch} (a) and (c), the accuracy on CIFAR10 increases with more epochs and increasing $I(z,v)$ cannot make a difference. But the accuracy on STL-10 decreases with more epochs, and increasing $I(z,v)$ can significantly improve the accuracy and does not decrease with more epochs. In fact, we use the \emph{unlabeled} split for contrastive training on STL-10, so it is intuitive that the shared information between views is not sufficient for the classification tasks on the \emph{train} and \emph{test} split. 

\paragraph{Increasing $I(z,x)$ in supervised learning.}
According to the information bottleneck theory \cite{tishby2015deep}, a model extracts the approximate minimal sufficient statistics of the input $x$ with respect to the label $y$ in supervised learning. In other words, the representation $z$ only contains the information related to the label and eliminates other irrelevant information which is considered as noise. However, label-irrelevant information may be useful for some downstream tasks, so we evaluate the effect of increasing $I(z,x)$ in supervised learning. We train the ResNet18 backbone using the cross-entropy classification loss on CIFAR10 and CIFAR100, and choose $\lambda_1=\lambda_2=\lambda$ from $\{0.001,0.01,0.1,1,10\}$. The linear evaluation results are shown in \cref{super} and the best result in each block is in bold. As we can see, increasing $I(z,x)$ improves the performance on the transfer datasets and achieves comparable results on the source dataset, which means it can effectively alleviate the over-fitting on the label information. This discovery helps to obtain more general representations in the field of supervised pre-training and we left it for the future work.

\section{Limitations}
Our work has the following limitations. 1) Based on our experimental observation, the assumption that non-shared task-relevant information cannot be ignored usually well holds for the cross-domain transfer tasks, but may not be satisfied for the tasks on the training dataset. 2) Increasing $I(z,v)$ can also introduce noise (task-irrelevant) information which may increase the data demand in the downstream tasks, so one may need to adjust the coefficients $\lambda_1$ and $\lambda_2$ to achieve effective trade-off for the different downstream tasks. 3) Due to limited computing resources, we cannot reproduce the best results of SimCLR on ImageNet which need the batch size of 4096 and more training epochs.

\section{Conclusions}
In this work, we explore the relationship between the learned representations and downstream tasks in contrastive learning. Although some works propose to learn the minimal sufficient representation, we theoretically and empirically verify that the minimal sufficient representation is not sufficient for downstream tasks because it loses non-shared task-relevant information. We find that contrastive learning approximately obtains the minimal sufficient representation, which means it may over-fit to the shared information between views. To this end, we propose to increase the mutual information between the representation and input to approximately introduce more non-shared task-relevant information when the downstream tasks are unknown. For the future work, we can consider combining the reconstruction models \cite{bao2021beit,he2021masked} and contrastive learning for convolutional neural networks or vision transformers, since reconstruction can learn more sufficient information and contrast can make the representations more discriminative.
% For the future work, we can consider the situation where the downstream task information is given. Then we can design task-customized views and objective terms which can directly introduce more task-relevant information.

%%%%%%%%% REFERENCES
{\small
\bibliographystyle{ieee_fullname}
\bibliography{egbib}
}

%%%%%%%% Appendix
\clearpage
\appendix

\section{Proofs of theorems}\label{proofs}
In this section, we provide the proofs of the theorems in the main text. Since the random variable $z_1=f_1(v_1)$ is the representation of random variable $v_1$ where $f_1$ is an encoding function, we have
\begin{assumption}\label{ass1}
Random variable $z_1$ is conditionally independent from any other variable $s$ in the system once random variable $v_1$ is observed, i.e., $I(z_1,s|v_1)=0, \forall s$.
\end{assumption}
This assumption is also adopted in \cite{DBLP:conf/iclr/Federici0FKA20}. When $f_1$ is a deterministic function, this assumption strictly holds. And when $f_1$ is a random function, the information in $z_1$ consists of the information from $v_1$ and the information introduced by the randomness of function $f_1$ which can be considered irrelevant to other variables in the system, so this assumption still holds. Next, we first present two lemmas for subsequent proofs.
\begin{lemma}\label{lem1}
Let $z_1^{suf}$ and $z_1^{min}$ are the sufficient representation and the minimal sufficient representation of view $v_1$ for $v_2$ in contrative learning respectively, we have
\begin{equation}\label{lem11}
    I(z_1^{min},v_2,T)=I(z_1^{suf},v_2,T)=I(v_1,v_2,T) 
\end{equation}
\begin{equation}\label{lem12}
    I(z_1^{min},T|v_2)=0
\end{equation}
\begin{proof}
1) From the Definition \ref{def1} and the \cref{ass1}, we have
\begin{align*}
    & I(v_1,v_2,T)-I(z_1^{suf},v_2,T) \\
  =\; & [I(v_1,v_2)-I(v_1,v_2|T)]-[I(z_1^{suf},v_2)-I(z_1^{suf},v_2|T)] \\
  =\; & I(z_1^{suf},v_2|T)-I(v_1,v_2|T) \\
  =\; & [H(v_2|T)-H(v_2|z_1^{suf},T)]-[H(v_2|T)-H(v_2|v_1,T)] \\
  =\; & H(v_2|v_1,T)-H(v_2|z_1^{suf},T) \\
  =\; & [I(z_1^{suf},v_2|v_1,T)+H(v_2|v_1,z_1^{suf},T)]\\ - & [I(v_1,v_2|z_1^{suf},T)+H(v_2|v_1,z_1^{suf},T)] \\
  =\; & I(z_1^{suf},v_2|v_1,T)-I(v_1,v_2|z_1^{suf},T) \\
  =\; & I(z_1^{suf},v_2|v_1,T)=0
\end{align*}
Therefore, we have 
\begin{equation*}
    I(z_1^{suf},v_2,T)=I(v_1,v_2,T)
\end{equation*}
The above proof process only uses the sufficiency of $z_1^{suf}$ for $v_2$, so we have
\begin{equation*}
    I(z_1^{min},v_2,T)=I(v_1,v_2,T)
\end{equation*}
2) From the Definition \ref{def2} and the \cref{ass1}, we have
\begin{equation*}
    I(z_1^{min},v_1|v_2)=0 \qquad I(z_1^{min},T|v_1)=0
\end{equation*}
Applying these two equations, we have
\begin{align*}
    I(z_1^{min},T|v_2) & =I(z_1^{min},T|v_1,v_2)+I(z_1^{min},T,v_1|v_2) \\
                       & =I(z_1^{min},T,v_1|v_2) \\
                       & =I(z_1^{min},v_1|v_2)-I(z_1^{min},v_1|T,v_2)=0
\end{align*}
\end{proof}
\end{lemma}
We consider the conditional entropy of the task variable $T$ given the representation $z_1$.
\begin{lemma}\label{lem2}
For arbitrary learned representation $z_1$, the conditional entropy $H(T|z_1)$ of the task variable $T$ given $z_1$ satisfies
\begin{equation}
    H(T|z_1)=H(T)-I(z_1,T|v_2)-I(z_1,v_2,T)
\end{equation}
Specifically, for the sufficient representation $z_1^{suf}$, the conditional entropy $H(T|z_1^{suf})$ satisfies
\begin{equation}
    H(T|z_1^{suf})=H(T)-I(z_1^{suf},T|v_2)-I(v_1,v_2,T)
\end{equation}
for the minimal sufficient representation $z_1^{min}$, the conditional entropy $H(T|z_1^{min})$ satisfies
\begin{equation}
    H(T|z_1^{min})=H(T)-I(v_1,v_2,T)
\end{equation}
\begin{proof}
We have
\begin{align*}
    H(T|z_1) & =H(T)-I(T,z_1) \\
             & =H(T)-[I(T,z_1,v_2)+I(T,z_1|v_2)] \\
             & =H(T)-I(z_1,T|v_2)-I(z_1,v_2,T)
\end{align*}
Applying the \cref{lem11}, the conditional entropy $H(T|z_1^{suf})$ satisfies
\begin{align*}
    H(T|z_1^{suf}) & =H(T)-I(z_1^{suf},T|v_2)-I(z_1^{suf},v_2,T)\\
                   & =H(T)-I(z_1^{suf},T|v_2)-I(v_1,v_2,T)
\end{align*}
Further, applying the \cref{lem12}, the conditional entropy $H(T|z_1^{min})$ satisfies
\begin{align*}
    H(T|z_1^{min}) & =H(T)-I(z_1^{min},T|v_2)-I(v_1,v_2,T) \\
                   & =H(T)-I(v_1,v_2,T)
\end{align*}
\end{proof}
\end{lemma}
Finally, we give the proofs of Theorem \ref{the1}, \ref{the2} and \ref{the3}.

\paragraph{The proof of Theorem \ref{the1}.}
\begin{proof}
We decompose the Theorem \ref{the1} into three equations and prove them in turn. \\
1) $I(v_1,T)=I(z_1^{min},T)+I(v_1,T|v_2)$.
\begin{align*}
    I(v_1,T) & =I(v_1,T,v_2)+I(v_1,T|v_2) \\
             & =I(z_1^{min},T,v_2)+I(v_1,T|v_2) \\
             & =I(z_1^{min},T)-I(z_1^{min},T|v_2)+I(v_1,T|v_2) \\
             & =I(z_1^{min},T)+I(v_1,T|v_2)
\end{align*}
2) $I(z_1^{suf},T)=I(z_1^{min},T)+I(z_1^{suf},T|v_2)$.
\begin{align*}
    I(z_1^{suf},T) & =I(z_1^{suf},T,v_2)+I(z_1^{suf},T|v_2) \\
             & =I(z_1^{min},T,v_2)+I(z_1^{suf},T|v_2) \\
             & =I(z_1^{min},T)-I(z_1^{min},T|v_2)+I(z_1^{suf},T|v_2) \\
             & =I(z_1^{min},T)+I(z_1^{suf},T|v_2)
\end{align*}
3) $I(v_1,T|v_2)\geq I(z_1^{suf},T|v_2)\geq 0$.

Applying the Data Processing Inequality \cite{DBLP:books/daglib/0016881} to the Markov chain $T\rightarrow v_1\rightarrow z_1^{suf}$, we have $I(v_1,T)\geq I(z_1^{suf},T)$, so
\begin{align*}
    I(v_1,T|v_2) & =I(v_1,T)-I(v_1,T,v_2) \\
                 & =I(v_1,T)-I(z_1^{suf},T,v_2) \\
                 &\geq I(z_1^{suf},T)-I(z_1^{suf},T,v_2)\\
                 &\geq I(z_1^{suf},T|v_2)\geq 0
\end{align*}
Combining these three equations, we can get Theorem \ref{the1}.
\end{proof}

\paragraph{The proof of Theorem \ref{the2}.}
\begin{proof}
According to \cite{feder1994relations}, the relationship between the Bayes error rate $P_e$ and the conditional entropy $H(T|z_1)$ is
\begin{equation*}
    -\ln(1-P_e)\leq H(T|z_1)
\end{equation*}
which is equivalent to
\begin{equation*}
    P_e\leq 1-\exp[-H(T|z_1)]
\end{equation*}
Applying the \cref{lem2}, for arbitrary learned representation $z_1$, its Bayes error rate $P_e$ satisfies
\begin{equation*}
P_e\leq 1-\exp[-(H(T)-I(z_1,T|v_2)-I(z_1,v_2,T))]
\end{equation*}
for the sufficient representation $z_1^{suf}$, its Bayes error rate $P_e^{suf}$ satisfies
\begin{equation*}
P_e^{suf}\leq 1-\exp[-(H(T)-I(z_1^{suf},T|v_2)-I(v_1,v_2,T))]
\end{equation*}
for the minimal sufficient representation $z_1^{min}$, its Bayes error rate $P_e^{min}$ satisfies
\begin{equation*}
P_e^{min}\leq 1-\exp[-(H(T)-I(v_1,v_2,T))]
\end{equation*}
Note that $0\leq P_e\leq 1-1/|T|$, so we use the threshold function $\Gamma(x)=\min\{\max\{x,0\},1-1/|T|\}$ to prevent overflow.
\end{proof}

\paragraph{The proof of Theorem \ref{the3}.}
\begin{proof}
According to \cite{frenay2013mutual}, when the conditional distribution $p(\varepsilon|z_1)$ of estimation error $\varepsilon$ is uniform, Laplace and Gaussian distribution, the minimum expected squared prediction error $R_e$ becomes $\frac{1}{12}\exp[2H(T|z_1)]$, $\frac{1}{2e^2}\exp[2H(T|z_1)]$ and $\frac{1}{2\pi e}\exp[2H(T|z_1)]$ respectively. Therefore, we unify them as
\begin{equation*}
    R_e=\alpha\cdot\exp[2H(T|z_1)]
\end{equation*}
where $\alpha$ is a constant coefficient which depends on the conditional distribution $p(\varepsilon|z_1)$. Applying the \cref{lem2}, for arbitrary learned representation $z_1$, we have
\begin{equation*}
R_e=\alpha\cdot\exp[2\cdot(H(T)-I(z_1,T|v_2)-I(z_1,v_2,T))]
\end{equation*}
for the sufficient representation $z_1^{suf}$, we have
\begin{equation*}
\resizebox{.95\linewidth}{!}{$R_e^{suf}=\alpha\cdot\exp[2\cdot(H(T)-I(z_1^{suf},T|v_2)-I(v_1,v_2,T))]$}
\end{equation*}
for the minimal sufficient representation $z_1^{min}$, we have
\begin{equation*}
R_e^{min}=\alpha\cdot\exp[2\cdot(H(T)-I(v_1,v_2,T))]
\end{equation*}
\end{proof}

\section{Choice of mutual information estimate}\label{Choice_LBE}
In our Implementation \uppercase\expandafter{\romannumeral2}, we need to use a mutual information lower bound estimate to calculate $I(z,v)$ where $v$ is the original input (e.g., images) and $z$ is the representation (feature vectors). We consider three candidate estimates:\\
1) The bound of Nguyen, Wainwright and Jordan \cite{nguyen2010estimating}
\begin{equation}
\hat{I}_{NWJ}(z,v)=\mathbb{E}_{p(z,v)}[h(z,v)]-\mathbb{E}_{p(z)p(v)}[e^{h(z,v)-1}]
\end{equation}
2) MINE \cite{belghazi2018mutual}
\begin{equation}
\resizebox{.88\linewidth}{!}{$\hat{I}_{MINE}(z,v)=\mathbb{E}_{p(z,v)}[h(z,v)]-\ln(\mathbb{E}_{p(z)p(v)}[e^{h(z,v)}])$}
\end{equation}
3) InfoNCE \cite{poole2019variational}
\begin{equation}
\hat{I}_{NCE}(z,v)=\mathbb{E}\left[\frac{1}{N}\sum_{k=1}^N\ln\frac{p(z^k|v^k)}{\frac{1}{N}\sum_{l=1}^Np(z^l|v^k)}\right]
\end{equation}
where $(z^k,v^k),k=1,\cdots,N$ are $N$ copies of $(z,v)$ and the expectation is over $\Pi_kp(z^k,v^k)$. As we can see, when we calculate the bound $\hat{I}_{NWJ}$ and $\hat{I}_{MINE}$, we need to calculate the critic $h(z,v)$ between the representation $z$ and original input $v$. If we use a neural network to model the critic $h(z,v)$, we have to take the original input (\eg images) and the representation together as the input of a neural network. Since the distribution of the original input $v$ and the representation $z$ is quite different, it is very difficult. Therefore, we use the InfoNCE lower bound estimate.

\section{More experiments}
In this section, we provide more experiments to support our work.

\begin{table*}
\centering
\scalebox{0.9}{
\begin{tabular}{l|c|cccccc}
\toprule[0.1em]
Model      & CIFAR10 & DTD & MNIST & FaMNIST & CUBirds & VGGFlower & TrafficSigns \\
\hline
BarTwins     & 86.85 & 28.56 & 95.39 & 86.19 & 7.49 & 35.91 & 88.50 \\
BarTwins+RC (ours)  & \textbf{86.91} & 28.97 & 96.60 & 86.72 & 7.90 & 38.94 & 90.92 \\
BarTwins+LBE (ours) & 86.38 & \textbf{29.54} & \textbf{96.72} & \textbf{86.88} & \textbf{8.47} & \textbf{41.44} & \textbf{92.76} \\
\toprule[0.1em]
Model      & STL-10 & DTD & MNIST & FaMNIST & CUBirds & VGGFlower & TrafficSigns \\
\hline
BarTwins       & 80.59 & 36.86 & 94.27 & 86.63 & 7.47 & 44.89 & 73.73 \\
BarTwins+RC (ours)  & \textbf{82.21} & 36.97 & 94.45 & 86.71 & 7.89 & 46.31 & 78.94 \\
BarTwins+LBE (ours) & 81.13 & \textbf{37.32} & \textbf{96.33} & \textbf{87.13} & \textbf{8.08} & \textbf{49.82} & \textbf{82.08} \\
\hline
\end{tabular}}
\caption{Linear evaluation accuracy ($\%$) on the source dataset (CIFAR10 or STL-10) and other transfer datasets.}
\label{bar_le}
\end{table*}

\subsection{Results on Barlow Twins}\label{bartwins}
In the main text, we provide the results on two classic contrastive learning models: SimCLR \cite{chen2020simple} and BYOL \cite{grill2020bootstrap}. SimCLR perfectly matches the contrastive learning framework, maximizing the lower bound estimate of the mutual information $I(z_1,z_2)$. BYOL avoids the dependence on the large amount of negative samples, and adopts prediction loss and the asymmetric structure. We further verify the effectiveness of increasing $I(z,v)$ on Barlow Twins \cite{DBLP:conf/icml/ZbontarJMLD21} which makes the cross-correlation matrix between the representations of different views as close to the identity matrix as possible. Although the loss functions of these contrastive learning models are very different, they all satisfy the internal mechanism that the views provide supervision information to each other, so they all approximately learn the minimal sufficient representation. We use the same pre-training schedule and linear evaluation protocol as in the main text and set $\lambda_1=\lambda_2=1$. For STL-10, we use the \emph{unlabeled} split for contrastive learning and the \emph{train} and \emph{test} split for linear evaluation.

The results are shown in \cref{bar_le} and the best result in each block is in bold. Increasing $I(z,v)$ can improve the accuracy of the learned representations in Barlow Twins in downstream classification tasks, which indicates that our analysis results are applicable to various contrastive losses.

\subsection{Reconstructed samples}
\begin{figure}
\begin{center}
\includegraphics[width=0.95\linewidth]{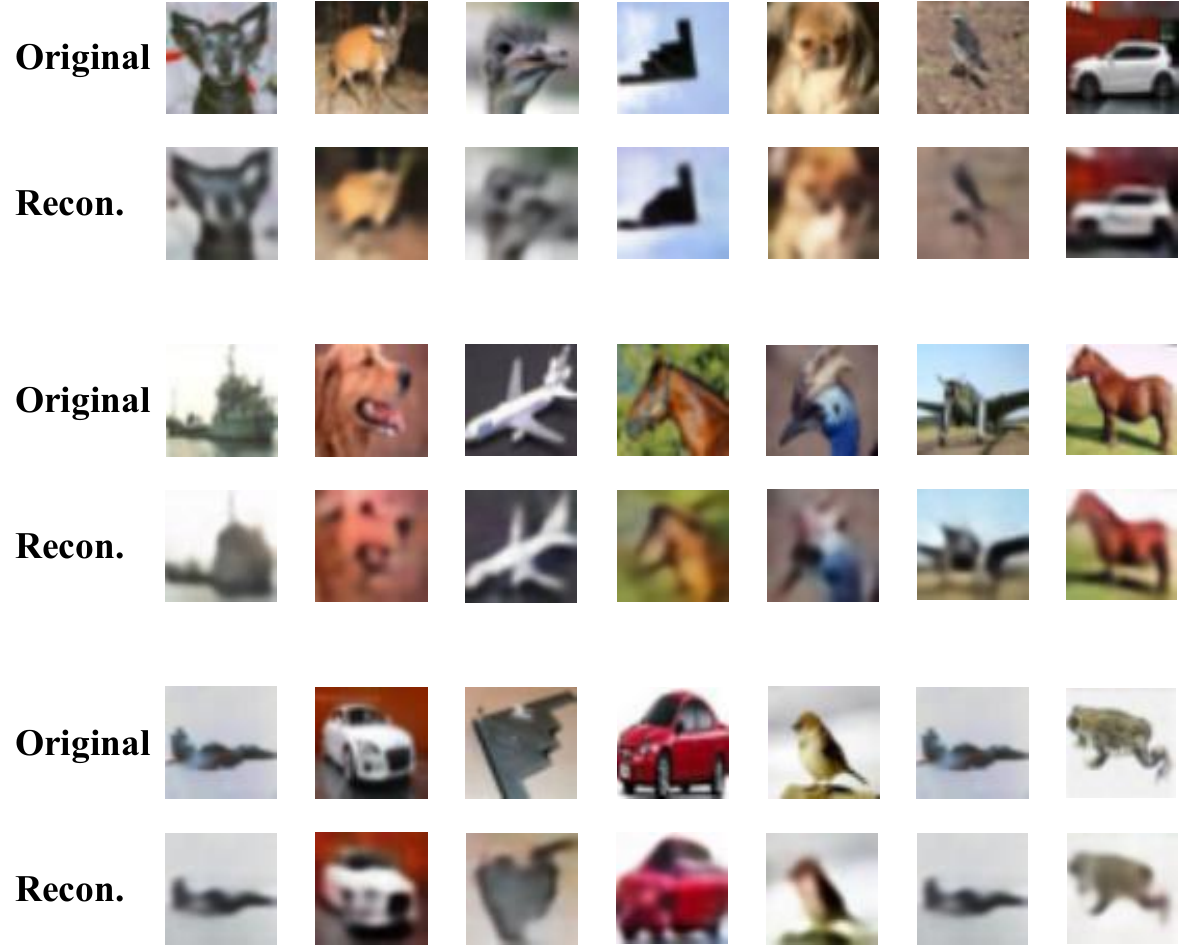}
\end{center}
\caption{Demonstration of the reconstruction effect of our Implementation \uppercase\expandafter{\romannumeral1}. We provide the original input images and the reconstructed images for comparison. We use SimCLR contrastive loss and take CIFAR10 as the training dataset.}
\label{recon}
\end{figure}

In order to show the reconstruction effect of our Implementation \uppercase\expandafter{\romannumeral1}, we provide the reconstructed images after training. As an example, we use SimCLR contrastive loss and take CIFAR10 as the training dataset. The original input images and the reconstructed images are shown in \cref{recon}. As we can see, the reconstructed images retain the shape and outline information in the original images, so as the obtained representations. Since we use the mean square error loss to optimize the reconstruction module, the reconstructed images are blurry and this phenomenon is also observed in vanilla variational auto-encoder \cite{DBLP:journals/corr/KingmaW13}.

\section{Derivation of $L_{MIB}$ and $L_{IP}$}\label{MIB_IP}
Federici \etal \cite{DBLP:conf/iclr/Federici0FKA20} and Tsai \etal \cite{DBLP:conf/iclr/Tsai0SM21} propose to eliminate the non-shared information between views in the representation to get the minimal sufficient representation. To this end, they propose their respective regularization terms. Here we derive the specific forms used in the main text.

In \cite{DBLP:conf/iclr/Federici0FKA20}, the regularization term is
\begin{align}
    L_{MIB} &=D_{SKL}(p(z_1|v_1)|| p(z_2|v_2)) \nonumber \\
            &=\frac{1}{2}[KL(p(z_1|v_1)||p(z_2|v_2)) \nonumber \\ & +KL(p(z_2|v_2)||p(z_1|v_1))]
\end{align}
According to the description in their paper and the official code \footnote{\url{https://github.com/mfederici/Multi-View-Information-Bottleneck}},
they model the two stochastic encoders $p(z_1|v_1)$ and $p(z_2|v_2)$ as 
\begin{align}
    p(z_1|v_1) &=\mathcal{N}(z_1;\mu_1,\text{diag}(\sigma_1^2)) \\
    p(z_2|v_2) &=\mathcal{N}(z_2;\mu_2,\text{diag}(\sigma_2^2))
\end{align}
where $\mu_1(v_1)$,$\sigma_1^2(v_1)$,$\mu_2(v_2)$ and $\sigma_2^2(v_2)$ are all functions of the input ($v_1$ or $v_2$), $\text{diag}(e)$ creates a matrix in which the diagonal elements consist of vector $e$ and all off-diagonal elements are zeros. The regularization term has the analytical expression
\begin{equation}
\resizebox{.9\linewidth}{!}{$L_{MIB}=\frac{1}{4}\sum_{i=1}^d\left[\frac{\sigma_1^{i2}}{\sigma_2^{i2}}+\frac{\sigma_2^{i2}}{\sigma_1^{i2}}+\frac{(\mu_1^i-\mu_2^i)^2}{\sigma_2^{i2}}+\frac{(\mu_2^i-\mu_1^i)^2}{\sigma_1^{i2}}-2\right]$}
\end{equation}
where $d$ is the dimension of $z_1$ and $z_2$. We want to minimize $L_{MIB}$, and when $\sigma_1^2=\sigma_2^2$, the term $\sigma_1^{i2}/\sigma_2^{i2}+\sigma_2^{i2}/\sigma_1^{i2}$ takes the minimum value $2$, so the regularization term becomes
\begin{equation}\label{MIB_dev}
    L_{MIB}=\frac{1}{2}\sum_{i=1}^d\frac{(\mu_1^i-\mu_2^i)^2}{\sigma_1^{i2}}
\end{equation}
In practice, minimizing $L_{MIB}$ makes the variance $\sigma_1^2$ and $\sigma_2^2$ very large, and the sampled representations change drastically and have very poor performance in downstream tasks. If the upper bound of the variance $\sigma_1^2$ and $\sigma_2^2$ is fixed, such as using the sigmoid activation function to limit it to $(0,1)$, they will converge to the maximum value as the training progresses. Therefore, we might as well fix the variance and model the two stochastic encoders $p(z_1|v_1)$ and $p(z_2|v_2)$ as
\begin{align}
    p(z_1|v_1) &=\mathcal{N}(z_1;f_1(v_1),\sigma^2I) \\
    p(z_2|v_2) &=\mathcal{N}(z_2;f_2(v_2),\sigma^2I)
\end{align}
where $I$ is the identity matrix, $\sigma^2$ is the given variance, $f_i,i=1,2$ are deterministic encoders. This also guarantees a fair comparison with our Implementation \uppercase\expandafter{\romannumeral2}. According to the \cref{MIB_dev}, the regularization term is equivalent to
\begin{equation}
    L_{MIB}=\|f_1(v_1)-f_2(v_2)\|_2^2
\end{equation}
We calculate the expectation of the regularization term on the data distribution $p(v_1,v_2)$ and get
\begin{equation}
    L_{MIB}=\mathbb{E}_{p(v_1,v_2)}[\|f_1(v_1)-f_2(v_2)\|_2^2]
\end{equation}

In \cite{DBLP:conf/iclr/Tsai0SM21}, the authors define the inverse predictive loss
\begin{equation}
\resizebox{.88\linewidth}{!}{$L_{IP}=\mathbb{E}_{p(v_1,v_2)}[\|z_1-z_2\|_2^2]=\mathbb{E}_{p(v_1,v_2)}[\|f_1(v_1)-f_2(v_2)\|_2^2]$}
\end{equation}

\end{document}